\documentclass{article} 
\usepackage{nips15submit_e,times}
\usepackage{hyperref}
\usepackage{url}

\usepackage{algorithm}
\usepackage{algorithmic}

\usepackage{epsf}
\usepackage{epsfig}
\usepackage{multirow}
\usepackage{tikz}
\usepackage{mathtools}
\usepackage{bm}
\usepackage{amsmath, amsthm, amssymb}
\usepackage{bbm}
\usepackage{dsfont}
\usepackage{wrapfig}
\usepackage{enumitem}

\DeclarePairedDelimiter\floor{\lfloor}{\rfloor}

\newcommand{\xdomain}{\ensuremath{\mathcal{X}}}
\newcommand{\R}{\ensuremath{\mathbb{R}}}
\renewcommand{\H}{\ensuremath{\mathcal{H}}}

\newcommand{\comment}[1]{}

\long\def\ignore#1{}

\newtheorem{theorem}{Theorem}
\newtheorem{lemma}{Lemma}

\newtheorem{definition}{Definition}

\title{Logarithmic Time Online Multiclass prediction}

\author{
Anna Choromanska \\
Courant Institute of Mathematical Sciences\\
New York, NY, USA\\
\texttt{achoroma@cims.nyu.edu} \\
\And
John Langford \\
Microsoft Research \\
New York, NY, USA \\
\texttt{jcl@microsoft.com} \\
}

\nipsfinalcopy 

\begin{document}

\maketitle

\begin{abstract} 
We study the problem of multiclass classification with an extremely
large number of classes ($k$), with the goal of obtaining train and test
time complexity logarithmic in the number of classes. We develop
top-down tree construction approaches for constructing logarithmic
depth trees. On the theoretical front, we formulate a new objective
function, which is optimized at each node of the tree and creates
dynamic partitions of the data which are both pure (in terms of class
labels) and balanced. We demonstrate that under favorable conditions,
we can construct logarithmic depth trees that have leaves with low
label entropy. However, the objective function at the nodes is
challenging to optimize computationally. We address the empirical
problem with a new online decision tree construction procedure.
Experiments demonstrate that this online algorithm quickly achieves
improvement in test error compared to more common logarithmic training time approaches, which makes it a plausible method in computationally constrained large-$k$ applications.
\end{abstract} 

\section{Introduction}
The central problem of this paper is computational complexity in a
setting where the number of classes $k$ for multiclass prediction is
very large.  Such problems occur in natural language (Which
translation is best?), search (What result is best?), and detection
(Who is that?) tasks.  Almost all machine learning algorithms (with
the exception of decision trees) have running times for
multiclass classification which are $\mathcal{O}(k)$ with a canonical
example being one-against-all classifiers~\cite{Rifkin2004}.  

In this setting, the most efficient possible accurate approach is
given by information theory~\cite{CnT}. In essence, any multiclass
classification algorithm must uniquely specify the bits of all labels
that it predicts correctly on.  Consequently, Kraft's inequality
(\cite{CnT} equation 5.6) implies that the expected
\emph{computational} complexity of predicting correctly is
$\Omega(H(Y))$ per example where $H(Y)$ is the Shannon entropy of the
label.  For the worst case distribution on $k$ classes, this implies
$\Omega(\log(k))$ computation is required. 

Hence, our goal is achieving $O(\log(k))$ computational time per example\footnote{Throughout the paper by logarithmic time we mean logarithmic time per example.} for both
training and testing, while effectively using online learning
algorithms to minimize passes over the data.

The goal of logarithmic (in $k$) complexity naturally motivates
approaches that construct a logarithmic depth hierarchy over the
labels, with one label per leaf. While this hierarchy is sometimes
available through prior knowledge, in many scenarios it needs to be
learned as well. This naturally leads to a \emph{partition} problem
which arises at each node in the hierarchy. The partition problem is
finding a classifier: $c:X \rightarrow \{-1,1\}$ which divides
examples into two subsets with a purer set of labels than the original
set. Definitions of purity vary, but canonical examples are the number
of labels remaining in each subset, or softer notions such as the
average Shannon entropy of the class labels. Despite resulting in a
classifier, this problem is fundamentally different from standard
binary classification. To see this, note that replacing $c(x)$ with
$-c(x)$ is very bad for binary classification, but has no impact on
the quality of a partition\footnote{The problem bears parallels to
  clustering in this regard.}. The partition problem is fundamentally
non-convex for symmetric classes since the average $\frac{c(x) -
  c(x)}{2}$ of $c(x)$ and $-c(x)$ is a poor partition (the always-$0$ function places all points on the same side).

The choice of partition matters in problem dependent ways.  For
example, consider examples on a line with label $i$ at position $i$
and threshold classifiers.  In this case, trying to partition class
labels $\{1,3\}$ from class label $2$ results in poor performance.

The partition problem is typically solved for decision tree learning
via an enumerate-and-test approach amongst a small set of possible
classifiers (see e.g.~\cite{ig}). In the multiclass setting, it is desirable to achieve
substantial error reduction for each node in the tree which motivates
using a richer set of classifiers in the nodes to minimize the number
of nodes, and thereby decrease the computational complexity. The main
theoretical contribution of this work is to establish a boosting
algorithm for learning trees with $O(k)$ nodes and $O(\log k)$ depth,
thereby addressing the goal of logarithmic time train and test
complexity. Our main theoretical result, presented in Section~\ref{sec:boosting}, generalizes a
binary boosting-by-decision-tree theorem~\cite{Kearns95} to multiclass
boosting. As in all boosting results, performance is critically
dependent on the quality of the \emph{weak learner}, supporting
intuition that we need sufficiently rich partitioners at nodes.  The
approach uses a new objective for decision tree learning, which we
optimize at each node of the tree.  The objective and its theoretical
properties are presented in Section~\ref{sec:framework}.
\vspace{-0.17in}
\begin{wrapfigure}{l}{0.5\textwidth} 
  \begin{center}
\includegraphics[scale = 0.35]{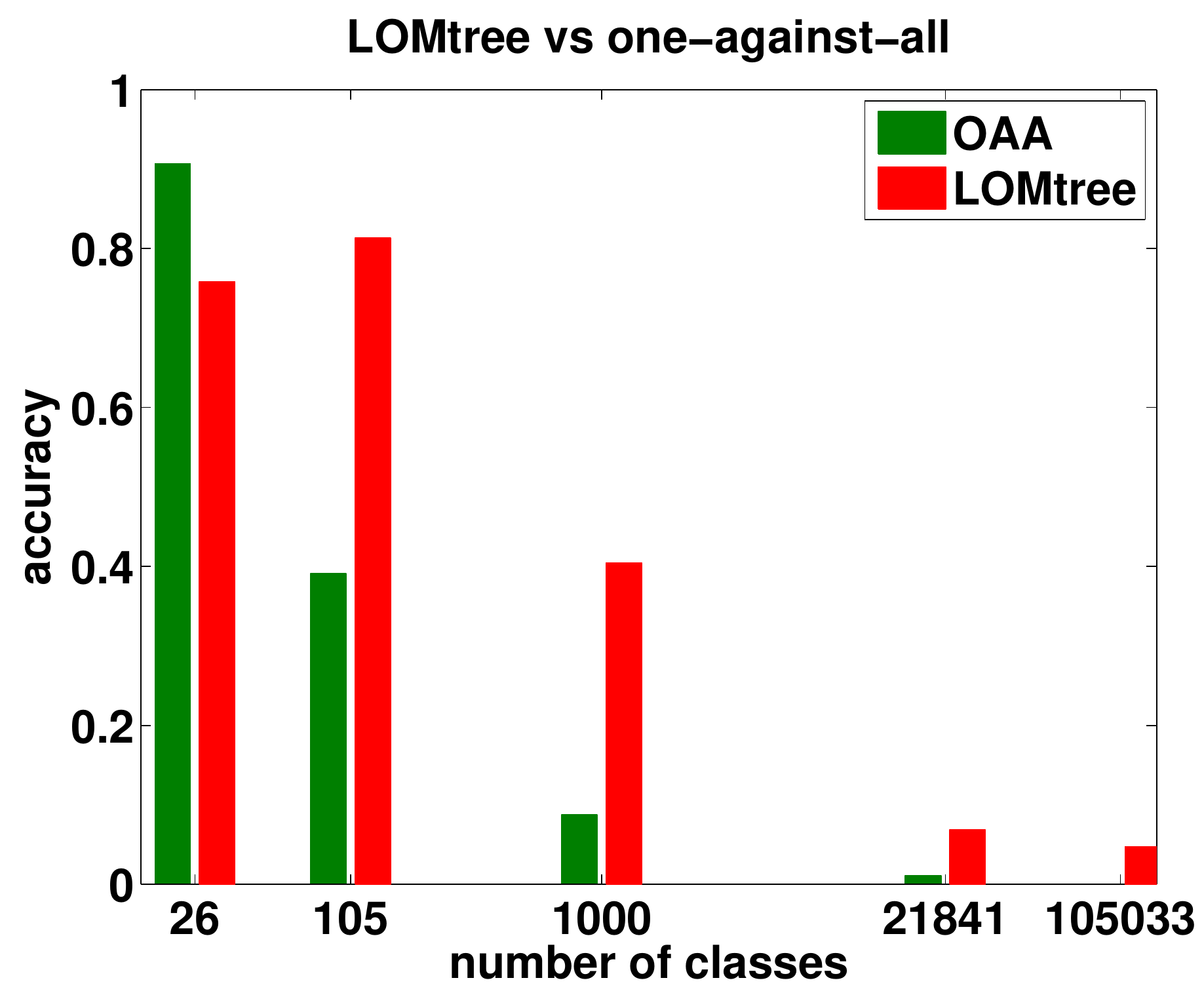}
\end{center}
\vspace{-0.22in}
\caption{A comparison of One-Against-All (OAA) and the Logarithmic Online
  Multiclass Tree (LOMtree) with One-Against-All constrained to use
  the same training time as the LOMtree by dataset truncation and
  LOMtree constrained to use the same representation complexity as
  One-Against-All.  As the number of class labels grows, the problem
  becomes harder and the LOMtree becomes more dominant.}
\vspace{-0.25in}
\end{wrapfigure}

A complete system with multiple partitions could be constructed top
down (as the boosting theorem) or bottom up (as Filter
tree~\cite{BeygelzimerLR09}).  A bottom up partition process appears
impossible with representational constraints as shown in Section~\ref{sec:bottom-up} in the Supplementary material so we focus on top-down tree creation.

Whenever there are representational constraints on partitions (such as
linear classifiers), finding a strong partition function requires an
efficient search over this set of classifiers. Efficient searches over
large function classes are routinely performed via gradient descent
techniques for supervised learning, so they seem like a natural
candidate. In existing literature, examples for doing this exist when
the problem is indeed binary, or when there is a prespecified
hierarchy over the labels and we just need to find partitioners
aligned with that hierarchy.
Neither of these cases applies---we have multiple labels and want to
dynamically create the choice of partition, rather than assuming that
one was handed to us. Does there exist a purity criterion amenable to
a gradient descent approach? The precise objective studied in theory
fails this test due to its discrete nature, and even natural
approximations are challenging to tractably optimize under
computational constraints. As a result, we use the theoretical
objective as a motivation and construct a new Logarithmic Online
Multiclass Tree (LOMtree) algorithm for empirical evaluation. 

Creating a tree in an online fashion creates a new class of problems.
What if some node is initially created but eventually proves useless
because no examples go to it?  At best this results in a wasteful
solution, while in practice it starves other parts of the tree which
need representational complexity.  To deal with this, we design an
efficient process for recycling orphan nodes into locations where they
are needed, and prove that the number of times a node is recycled is
at most logarithmic in the number of examples.  The algorithm is
described in Section~\ref{sec:alg} and analyzed in
Section~\ref{sec:swap-bound}.

And is it effective? Given the inherent non-convexity of the partition
problem this is unavoidably an empirical question which we answer on a
range of datasets varying from 26 to 105K classes in
Section~\ref{sec:experiments}.  We find that under constrained
training times, this approach is quite effective compared to all
baselines while dominating other $O(\log k)$ train time approaches.

What's new? To the best of our knowledge, the splitting criterion, the
boosting statement, the LOMtree algorithm, the swapping guarantee,
and the experimental results are all new here.

\subsection{Prior Work}

Only a few authors address logarithmic time
training. The Filter tree~\cite{BeygelzimerLR09} addresses consistent (and
robust) multiclass classification, showing that it is possible in the
statistical limit. The Filter tree does not address the partition
problem as we do here which as shown in our experimental section is
often helpful. The partition finding problem is addressed in the conditional
probability tree~\cite{BeygelzimerLLSS09}, but that paper addresses
conditional probability estimation. Conditional probability
estimation can be converted into multiclass prediction~\cite{Bishop:2006:PRM:1162264}, but doing so
is not a logarithmic time operation.

Quite a few authors have addressed logarithmic testing time while
allowing training time to be $O(k)$ or worse. While these approaches
are intractable on our larger scale problems, we describe them here
for context. The partition problem can be addressed by recursively applying
spectral clustering on a confusion graph~\cite{BengioWG10} (other clustering approaches include~\cite{journals/informaticaSI/MadzarovGC09}).
Empirically, this approach has been found to sometimes lead to badly
imbalanced splits~\cite{DengSBL11}. In the context of ranking, another approach uses $k$-means
hierarchical clustering to recover the label sets for a given
partition~\cite{weston13}. 

The more recent work~\cite{conf/cvpr/ZhaoX13} on the multiclass
classification problem addresses it via sparse output coding by tuning high-cardinality multiclass categorization into a
bit-by-bit decoding problem.  The authors decouple
the learning processes of coding matrix and bit predictors
and use probabilistic decoding to decode the optimal class
label. The authors however specify a class similarity which is $\mathcal{O}(k^2)$ to compute (see Section $2.1.1$ in~\cite{conf/cvpr/ZhaoX13}), and hence this approach is in a different complexity class than ours (this is also born out experimentally). The variant of the popular error correcting output code scheme for solving multi-label prediction problems with large output spaces under the assumption of output sparsity was also considered in~\cite{DBLP:journals/corr/abs-0902-1284}. Their approach in general requires $O(k)$ running time to decode since, in essence, the fit of each label to the predictions must be checked and there are $\mathcal{O}(k)$ labels. Another approach~\cite{DBLP:journals/corr/AgarwalKKSV13} proposes iterative least-squares-style algorithms for multi-class (and multi-label) prediction with relatively large number of examples
and data dimensions, and the work of~\cite{icml2014c2_beijbom14} focusing in particular on the cost-sensitive multiclass classification. Both approaches however have $\mathcal{O}(k)$ training time.

Decision trees are naturally structured to allow logarithmic time
prediction. Traditional decision trees often have difficulties with a
large number of classes because their splitting criteria are not
well-suited to the large class setting. However, newer
approaches~\cite{Manik,Prabhu2014} have addressed this effectively at
significant scales in the context of multilabel classification (multilabel learning, with missing labels, is also addressed in~\cite{LMLML14}). More specifically, the first work~\cite{Manik} performs brute force optimization of a multilabel variant of the Gini index defined over the set of positive labels in the node and assumes label independence during random forest construction. Their method makes fast predictions, however has high training costs~\cite{Prabhu2014}. The second work~\cite{Prabhu2014} optimizes a rank sensitive loss function (Discounted Cumulative Gain). Additionally, a well-known problem with hierarchical classification is
that the performance significantly deteriorates lower in the
hierarchy~\cite{LiuLargeScale2005} which some authors solve by biasing
the training distribution to reduce error propagation while
simultaneously combining bottom-up and top-down approaches during training~\cite{conf/sigir/BennettN09}.

The reduction approach we use for optimizing partitions implicitly
optimizes a differential objective. A non-reductive approach to this
has been tried previously~\cite{MTSWIMC} on other objectives yielding
good results in a different context. 

\comment{
John: The below isn't even log time.  Why do we discuss it?

rather than the multiclass classification problem, and yet another
related work by~\cite{LMLML14} addresses the problem of multi-label
learning with missing labels by studying the empirical risk
minimization framework with a low-rank constraint. 

John: So tangential that we shouldn't discuss?

Finally, in the context of object detection, the recent work
by~\cite{40814} demonstrated the ability to efficiently scale up
existing detection systems to one hundred thousand object classes
though again the problem they address quite differs from the central
focus of this paper.
}

\section{Framework and theoretical analysis}
\label{sec:framework}

In this section we describe the essential elements of the approach,
and outline the theoretical properties of the resulting framework. We
begin with high-level ideas.

\subsection{Setting}
\label{sec:outline}

We employ a hierarchical approach for learning a multiclass decision
tree structure, training this structure in a \emph{top-down}
fashion. We assume that we receive examples $x \in \xdomain \subseteq
\R^d$, with labels $y \in \{1,2,\ldots, k\}$. We also assume access to
a hypothesis class $\H$ where each $h \in \H$ is a binary classifier,
$h~:~\xdomain\mapsto \{-1, 1\}$. The overall objective is to learn a
tree of depth $O(\log k)$, where each node in the tree consists of a
classifier from $\H$. The classifiers are trained in such a way that
$h_n(x) = 1$ ($h_n$ denotes the classifier in node $n$ of the tree\footnote{Further in the paper we skip index $n$ whenever it is clear from the context that we consider a fixed tree node.}) means that the example $x$ is sent to the right subtree
of node $n$, while $h_n(x) = -1$ sends $x$ to the left subtree. When we
reach a leaf, we predict according to the label with the
highest frequency amongst the examples reaching that leaf.

In the interest of computational complexity, we want to encourage the
number of examples going to the left and right to be \emph{fairly
  balanced}. For good statistical accuracy, we want to send examples
of class $i$ almost exclusively to either the left or the right
subtree, thereby refining the \emph{purity} of the class distributions
at subsequent levels in the tree. The \emph{purity} of a tree node is therefore a measure of whether the examples of each class reaching the node are then mostly sent to its one child node (pure split) or otherwise to both children (impure split). The formal definitions of \textit{balancedness} and \textit{purity} are introduced in Section~\ref{sec:objective}. An objective expressing both criteria\footnote{We want an objective to achieve its optimum for simultaneously pure and balanced split.  The standard entropy-based criteria, such as Shannon or Gini entropy, as well as the criterion we will propose, posed in Equation~\ref{eqn:objective}, satisfy this requirement (for the entropy-based criteria see~\cite{Kearns95}, for our criterion see Lemma~\ref{lemma:maximal}).} and resulting theoretical properties are illustrated in the following sections. A key consideration in picking
this objective is that we want to effectively optimize it over
hypotheses $h \in \H$, while streaming over examples in an online
fashion\footnote{Our algorithm could also be implemented as batch or streaming, where in case of the latter one can for example make one pass through the data per every tree level, however for massive datasets making multiple passes through the data is computationally costly, further justifying the need for an online approach.}. This seems unsuitable with some of the more standard decision
tree objectives such as Shannon or Gini entropy, which leads us to
design a new objective. At the same time, we show in
Section~\ref{sec:boosting} that under suitable assumptions, optimizing
the objective also leads to effective reduction of the average Shannon
entropy over the entire tree.

\subsection{An objective and analysis of resulting partitions}
\label{sec:objective}

We now define a criterion to measure the quality of a hypothesis
$h \in \H$ in creating partitions at a fixed node $n$ in the tree. Let $\pi_i$ denotes the proportion of label $i$
amongst the examples reaching this node. Let $P(h(x) > 0)$ and $P(h(x)
> 0 | i)$ denote the fraction of examples reaching $n$ for which $h(x)
> 0$, marginally and conditional on class $i$ respectively. Then we
define the objective\footnote{The proposed objective function exhibits some similarities with the so-called Carnap's measure~\cite{Tentori2007107,Carnap1962} used in probability and inductive logic.}:
\vspace{-0.05in}
\begin{equation}
  J(h) = 2\sum_{i=1}^k \pi_i \left| P(h(x) > 0) - P(h(x) > 0 | i)
  \right|.
  \label{eqn:objective}
\vspace{-0.03in}
\end{equation} 
We aim to \emph{maximize the objective $J(h)$} to obtain
high quality partitions. Intuitively, the objective encourages the
fraction of examples going to the right from class $i$ to be
substantially different from the background fraction for each class
$i$. As a concrete simple scenario, if $P(h(x) > 0) = 0.5$ for some
hypothesis $h$, then the objective prefers $P(h(x) > 0 | i)$ to
be as close to 0 or 1 as possible for each class $i$, leading to pure
partitions. We now make these intuitions more formal.

\begin{definition}[Purity]
The hypothesis $h \in \mathcal{H}$ induces a pure split if
\vspace{-0.03in}
\[\alpha := \sum_{i=1}^{k}\pi_i\min(P(h(x) > 0|i), P(h(x)<0|i)) \leq
\delta, 
\vspace{-0.04in}
\]
where $\delta \in [0,0.5)$, and $\alpha$ is called the
  \emph{purity factor}.
\end{definition}
In particular, a partition is called \emph{maximally pure} if $\alpha
= 0$, meaning that each class is sent exclusively to the left or the
right. We now define a similar
definition for the balancedness of a split.

\begin{definition}[Balancedness]
The hypothesis $h \in \mathcal{H}$ induces a balanced split if
\vspace{-0.02in}
\[c \leq \underbrace{P(h(x) > 0)}_{=\beta} \leq 1 - c,
\vspace{-0.06in}
\]
where $c \in (0,0.5]$, and $\beta$ is called the \emph{balancing
    factor}.
\end{definition}
A partition is called \emph{maximally balanced} if $\beta = 0.5$,
meaning that an equal number of examples are sent to the left and
right children of the partition. The balancing factor and the purity factor are related as shown in Lemma~\ref{lemma:obj-to-purity} (the proofs of Lemma~\ref{lemma:obj-to-purity} and the following lemma (Lemma~\ref{lemma:maximal}) are deferred to the Supplementary material).

\begin{lemma}
  For any hypothesis $h$, and any distribution over examples $(x,y)$,
  the purity factor $\alpha$ and the balancing factor $\beta$ satisfy $\alpha \leq \min\{(2 - J(h))/(4\beta) - \beta, 0.5\}$.
\label{lemma:obj-to-purity}
\end{lemma}
A partition is called \emph{maximally pure and balanced} if it
satisfies both $\alpha = 0$ and $\beta = 0.5$.  We see that $J(h) = 1$
for a hypothesis $h$ inducing a maximally pure and balanced partition
as captured in the next lemma. Of course we do not expect to have
hypotheses producing maximally pure and balanced splits in practice.

\begin{lemma}
  For any hypothesis $h~:~\xdomain \mapsto \{-1,1\}$, the objective
  $J(h)$ satisfies $J(h) \in
  [0,1]$. Furthermore, if $h$ induces a
  maximally pure and balanced partition then $J(h) = 1$.
  \label{lemma:maximal}
\end{lemma}

\subsection{Quality of the entire tree}
\label{sec:boosting}

The above section helps us understand the quality of an individual
split produced by effectively maximizing $J(h)$. We next reason about the quality of the entire
tree as we add more and more nodes. We
measure the quality of trees using the average entropy over all the
leaves in the tree, and track the decrease of this entropy as a
function of the number of nodes. Our analysis extends the
theoretical analysis in~\cite{Kearns95}, originally developed to show the boosting
properties of the decision trees for binary classification
problems, to the multiclass
classification setting.

Given a tree $\mathcal{T}$, we consider the entropy function $G_t$ as the
measure of the quality of tree:
\vspace{-0.1in}
\[G_t = \sum_{l \in \mathcal{L}}w_l\sum_{i = 1}^k \pi_{l,i}\ln \left( \frac{1}{\pi_{l,i}} \right)
\vspace{-0.05in}
\]
where $\pi_{l,i}$'s are the probabilities that a randomly chosen data point $x$
drawn from $\mathcal{P}$, where $\mathcal{P}$ is a fixed target distribution over $\mathcal{X}$, has label
$i$ given that $x$ reaches node $l$, $\mathcal{L}$ denotes the set of all tree leaves, $t$ denotes the number of internal tree nodes, and $w_l$ is the weight of leaf $l$ defined as the probability a randomly chosen $x$ drawn from
$\mathcal{P}$ reaches leaf $l$ (note that $\sum_{l \in
  \mathcal{L}}w_l = 1$).

We next state the main theoretical result of this paper (it is captured in Theorem~\ref{thm:main}). We adopt the \emph{weak learning} framework. The \emph{weak hypothesis assumption}, captured in Definition~\ref{def:wha}, posits that each node of the tree $\mathcal{T}$ has a
hypothesis $h$ in its hypothesis class $\mathcal{H}$  which guarantees
simultaneously a "weak'' purity and a "weak'' balancedness of the split on any distribution $\mathcal{P}$ over
$\mathcal{X}$. Under this assumption, one can use the new decision
tree approach to drive the error below any threshold.

\begin{definition}[Weak Hypothesis Assumption]
Let $m$ denote any node of the tree
$\mathcal{T}$, and let $\beta_m = P(h_m(x) > 0)$ and $P_{m,i} = P(h_m(x) > 0|i)$. Furthermore, let $\gamma \in \mathbb{R}^{+}$ be such that for all $m$, $\gamma \in (0,\min(\beta_m,1-\beta_m)]$. We say that the \emph{weak hypothesis assumption} is satisfied when for any distribution
$\mathcal{P}$ over $\mathcal{X}$ at each node $m$ of the tree
$\mathcal{T}$ there exists a hypothesis $h_m \in \mathcal{H}$ such that $J(h_m)/2 =
\sum_{i = 1}^k \pi_{m,i}|P_{m,i} - \beta_{m}| \geq \gamma$.
\label{def:wha}
\end{definition}

\begin{theorem}
Under the Weak Hypothesis Assumption, for any $\alpha \in [0,1]$, to
obtain $G_t \leq \alpha$ it suffices to make $t \geq (1/\alpha)^\frac{4(1 - \gamma)^2\ln
  k}{\gamma^2}$ splits.
\label{thm:main}
\vspace{-0.05in}
\end{theorem}

We defer the proof of Theorem~\ref{thm:main} to the Supplementary material and provide its sketch now. The analysis studies a tree construction algorithm where we recursively find the leaf node with the highest weight, and choose to
split it into two children. Let $n$ be the
heaviest leaf at time $t$. Consider splitting it to two children. The contribution of node $n$ to
the tree entropy changes after it splits. This change (entropy reduction) corresponds to a gap
in the Jensen's inequality applied to the concave function, and thus can further be lower-bounded (we use the fact that Shannon entropy is strongly concave with respect to $\ell_1$-norm (see e.g., Example 2.5 in
Shalev-Shwartz~\cite{ShaiSS2012})). The obtained lower-bound turns out to depend proportionally on $J(h_n)^2$. This implies that the larger the objective
$J(h_n)$ is at time $t$, the larger the entropy reduction ends up being,
which further reinforces intuitions to maximize $J$. In general, it
might not be possible to find any hypothesis with a large enough
objective $J(h_n)$ to guarantee sufficient progress at this point so we
appeal to a \emph{weak learning assumption}. This assumption can be used to further lower-bound the entropy reduction and prove Theorem~\ref{thm:main}. 

\section{The LOMtree Algorithm}
\label{sec:alg}

\begin{algorithm}[h!]
\begin{tabular}{l}
\textbf{Input}: \:\:regression algorithm $R$, max number of tree non-leaf nodes $T$, swap resistance $R_S\:\:\:\:\:\:\:\:\:$\\
\hline
\hline
Subroutine \textbf{SetNode ($v$)}\\
\hline
  ${\bm m}_v = \emptyset $\:\:\:\:\!\:\!(${\bm m}_v(y)$ - sum of the scores for class $y$)\\
  \:${\bm l}_v \:\:= \emptyset $ \:\:\:\:\!\!(${\bm l}_v(y)$ - number of points of class $y$ reaching $v$)\\
  \:\:\!\!${\bm n}_v \:= \emptyset $ \:\:\!\:\:\!(${\bm n}_v(y)$ - number of points of class $y$ which are used to train regressor in $v$)\\
  \:${\bm e}_v \:\:\!= \emptyset $ \:\:\!\:\:\!(${\bm e}_v(y)$ - expected score for class $y$)\\
  $\:\:\!\!{\bm E}_v \:\!\:\!= 0 $ \:\:\!\:\:\!(expected total score)\\
  \:$C_v \:\:\!\!= 0$\:\:\:\:\:\!\!(the size of the smallest leaf\footnotemark[7] in the subtree with root $v$)\\
\hline
\hline
Subroutine \textbf{UpdateC} ($v$)\\
\hline
\textbf{While} ($v \neq r$ AND $C_{\textsc{parent}(v)} \neq C_v$)\\
\:\:\:\:\:\:\:$v = \textsc{parent}(v)$; \:\:\:$C_v = \min(C_{\textsc{left}(v)},C_{\textsc{right}(v)})$\footnotemark[8]\\
\hline
\hline
Subroutine \textbf{Swap} (v)\\
\hline
Find a leaf $s$ for which $(C_s = C_r)$\\
$s_{\textsc{pa}} \!\!=\!\! \textsc{parent}(s)$; $s_{\textsc{gpa}}$ \!\!=\! \textsc{grandpa}(s); $s_{\textsc{sib}} \!\!=\!\! \textsc{sibling}(s)$\footnotemark[9]\\ 
\textbf{If} ($s_{\textsc{pa}}$ = \textsc{left}($s_{\textsc{gpa}}$))\: \textsc{left}($s_{\textsc{gpa}}$) \!\:\:\:= $s_{\textsc{sib}}$\:\:\:\:\:\textbf{Else}\: \textsc{right}($s_{\textsc{gpa}}$) = $s_{\textsc{sib}}$\\
\textbf{UpdateC} ($s_{\textsc{sib}}$); \:\:\:\textbf{SetNode} ($s$); \:\:\:$\textsc{left}(v) = s$; \:\:\:\textbf{SetNode} ($s_{\textsc{pa}}$); \:\:\:$\textsc{right}(v) = s_{\textsc{pa}}$\\
\hline
\hline
\textbf{Create} root $r = 0$: \textbf{SetNode} ($r$); \:\:\:$t = 1$\\
\textbf{For each} example $(\bm{x},y)$ \textbf{do}\\
\:\:\:\:\:\:\:Set $j = r$\\
\:\:\:\:\:\:\:\textbf{Do}\\
\:\:\:\:\:\:\:\:\:\:\:\:\:\:\textbf{If} ($l_j(y) = \emptyset$)\\
\:\:\:\:\:\:\:\:\:\:\:\:\:\:\:\:\:\:\:\:$m_j(y) = 0$; \:\:\:$l_j(y) = 0$;\:\:\:$n_j(y) = 0$;\:\:\:$e_j(y) = 0$\\
\:\:\:\:\:\:\:\:\:\:\:\:\:\:$\bm{l}_j(y)$++\\
\:\:\:\:\:\:\:\:\:\:\:\:\:\:\textbf{If}($j$ is a leaf)\\
\:\:\:\:\:\:\:\:\:\:\:\:\:\:\:\:\:\:\:\:\textbf{If}(${\bm l}_j$ has at least $2$ non-zero entries)\\
\:\:\:\:\:\:\:\:\:\:\:\:\:\:\:\:\:\:\:\:\:\:\:\:\:\:\textbf{If}($t \!\!<\!\! T$ OR $C_j \!\!-\!\! \max_i{\bm l}_j(i) \!\!>\!\! R_S(C_r\!\!+\!\!1)$)\\
\:\:\:\:\:\:\:\:\:\:\:\:\:\:\:\:\:\:\:\:\:\:\:\:\:\:\:\:\:\:\:\:\textbf{If} ($t \!\!<\!\! T$)\\
\:\:\:\:\:\:\:\:\:\:\:\:\:\:\:\:\:\:\:\:\:\:\:\:\:\:\:\:\:\:\:\:\:\:\:\:\:\:\textbf{SetNode} (\textsc{left}($j$));\:\:\:\textbf{SetNode} (\textsc{right}($j$));\:\:\:$t$++\\
\:\:\:\:\:\:\:\:\:\:\:\:\:\:\:\:\:\:\:\:\:\:\:\:\:\:\:\:\:\:\:\:\textbf{Else}\: \textbf{Swap}(j)\\
\:\:\:\:\:\:\:\:\:\:\:\:\:\:\:\:\:\:\:\:\:\:\:\:\:\:\:\:\:\:\:\:$C_{\textsc{left}(j)} \!\!=\!\! \floor*{C_j/2}$;\:\:\:$C_{\textsc{right}(j)} \!\!=\!\! C_j \!\!-\!\! C_{\textsc{left}(j)}$;\:\:\:\textbf{UpdateC} (\textsc{left}($j$))\\
\:\:\:\:\:\:\:\:\:\:\:\:\:\:\textbf{If}($j$ is not a leaf)\\
\:\:\:\:\:\:\:\:\:\:\:\:\:\:\:\:\:\:\:\:\textbf{If} $\left(E_j > \bm{e}_j(y)\right)$\: $c \!=\! -1$\:\:\:\:\:\textbf{Else}\: $c \!=\! 1$\\
\:\:\:\:\:\:\:\:\:\:\:\:\:\:\:\:\:\:\:\:\textbf{Train} $h_j$ with example $({\bm x},c)$: $R({\bm x},c)$\\
\:\:\:\:\:\:\:\:\:\:\:\:\:\:\:\:\:\:\:\:$\bm{n}_j(y)+\!\!+$;\:\:\:${\bm m}_j(y)\:+\!\!= h_j(\bm{x})$;\:\:\:$\bm{e}_j(y) =\bm{m}_j(y)/\bm{n}_j(y)$;\:\:\:$E_j = \frac{\sum_{i = 1}^k{{\bm m}_j(i)}}{\sum_{i=1}^k{\bm n}_j(i)}$\footnotemark[10]\\
\:\:\:\:\:\:\:\:\:\:\:\:\:\:\:\:\:\:\:\:\textbf{Set} $j$ to the child of $j$ corresponding to $h_j$\\
\:\:\:\:\:\:\:\:\:\:\:\:\:\:\textbf{Else}\\
\:\:\:\:\:\:\:\:\:\:\:\:\:\:\:\:\:\:\:\:$C_j$++\\
\:\:\:\:\:\:\:\:\:\:\:\:\:\:\:\:\:\:\:\:\textbf{break}\\
\end{tabular}
\caption{LOMtree algorithm (online tree training)}
\label{alg:OTT}
\end{algorithm}

The objective function of Section~\ref{sec:framework} has another convenient form which yields a simple online algorithm for tree construction and training.  Note that Equation~\ref{eqn:objective} can be written (details are shown in Section~\ref{sec:explanation} in the Supplementary material) as
\vspace{-0.05in}
\[J(h) = 2\mathbb{E}_i [|\mathbb{E}_x[\mathds{1}(h(x) > 0)] - \mathbb{E}_{x}[\mathds{1}(h(x) > 0|i)]|].
\vspace{-0.05in}
\]
Maximizing this objective is a discrete optimization problem that can be relaxed as follows
\vspace{-0.05in}
\[J(h) = 2\mathbb{E}_i[|\mathbb{E}_x[h(x)] - \mathbb{E}_{x}[h(x)|i]|],
\vspace{-0.05in}
\]
where $E_{x}[h(x)|i]$ is the expected score of class i. 

We next explain our empirical approach for maximizing the relaxed objective. The empirical estimates of the expectations can be easily stored and updated online in every tree node. The decision whether to send an example reaching a node to its left or right child node is based on the sign of the difference between the two expectations: $\mathbb{E}_x[h(x)]$ and $\mathbb{E}_{x}[h(x)|y]$, where $y$ is a label of the data point, i.e. when $\mathbb{E}_x[h(x)] - \mathbb{E}_{x}[h(x)|y] > 0$ the data point is sent to the left, else it is sent to the right. This procedure is conveniently demonstrated on a toy example in Section~\ref{sec:toy} in the Supplement.

During training, the algorithm assigns a unique label to each node of the tree which is currently a leaf. This is the label with the highest frequency amongst the examples reaching that leaf. While testing, a test example is pushed down the tree along the path from the root to the leaf, where in each non-leaf node of the path its regressor directs the example either to the left or right child node. The test example is then labeled with the label assigned to the leaf that this example descended to.

The training algorithm is detailed in Algorithm~\ref{alg:OTT} where each tree node contains a classifier (we use linear classifiers), i.e. $h_j$ is the regressor stored in node $j$ and $h_j(\bf x)$ is the value of the prediction of $h_j$ on example $\bf x$\footnotetext[7]{The smallest leaf is the one with the smallest total number of data points reaching it in the past.}\footnotetext[8]{\textsc{parent}(v), \textsc{left}(v) and \textsc{right}(v) denote resp. the parent, and the left and right child of node $v$.}\footnotetext[9]{\textsc{grandpa}(v) and \textsc{sibling}(v) denote respectively the grandparent of node $v$ and the sibling of node $v$, i.e. the node which has the same parent as $v$.}\footnotetext[10]{In the implementation both sums are stored as variables thus updating $E_v$ takes $\mathcal{O}(1)$ computations.}\footnote[11]{We also refer to this prediction value as the 'score' in this section.}. The stopping criterion for expanding the tree is when the number of non-leaf nodes reaches a threshold $T$. 

\subsection{Swapping}
\label{sec:swap-bound}

\vspace{-0.1in}
\begin{figure}[htp!]
\center
\begin{minipage}{.5\textwidth}
\begin{tikzpicture}[level distance=0.6cm,
  level 1/.style={sibling distance=2.5cm},
  level 2/.style={sibling distance=1.25cm},scale=0.8]
  \node {$r$}
    child {node {$\dots$}
      child {node {$j$}}
      child {node {$\dots$}}
            }
    child {node {$\dots$}
                  child {node {\dots}}
                  child {node {$s_{\textsc{gpa}}$}
                       child {node {$\dots$}}
                       child {node {$s_{\textsc{pa}}$}
                            child {node {$s$}}
                            child {node {$s_{\textsc{sib}}$}
                                  child {node {$\dots$}}
				  child {node {$\dots$}}
                                    }
                               }
                          }
            };
\end{tikzpicture}
\end{minipage}%
\hspace{-0.7in}\begin{minipage}{.5\textwidth}
\centering
\begin{tikzpicture}[level distance=0.65cm,
  level 1/.style={sibling distance=2.5cm},
  level 2/.style={sibling distance=1.25cm},scale=0.8]
  \node {$r$}
    child {node {$\dots$}
      child {node {$j$}
        child {node {$s$}}
        child {node {$s_{\textsc{pa}}$}}
              }
      child {node {$\dots$}}
            }
    child {node {$\dots$}
                  child {node {\dots}}
                  child {node {$s_{\textsc{gpa}}$}
                       child {node {$\dots$}}
                       child {node {$s_{\textsc{sib}}$}
                            child {node {$\dots$}}
                            child {node {$\dots$}}
                               }
                          }
            };
\end{tikzpicture}
\end{minipage}%
\label{fig:swap}
\vspace{-0.1in}
\caption{Illustration of the swapping procedure. \textbf{Left:} before the swap, \textbf{right:} after the swap.}
\vspace{-0.2in}
\end{figure}
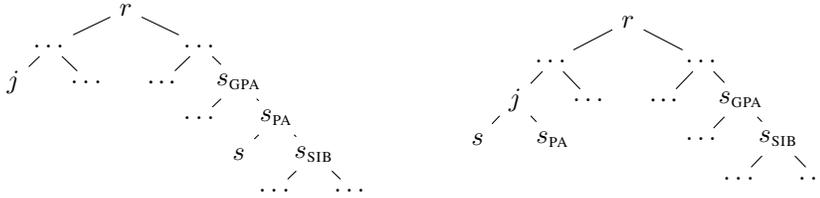

Consider a scenario where the current training example descends to
leaf $j$. The leaf can split (create two children) if the examples
that reached it in the past were coming from at least two different
classes. However, if the number of non-leaf nodes of the tree reaches
threshold $T$, no more nodes can be expanded and thus $j$ cannot
create children. Since the tree construction is done online, some
nodes created at early stages of training may end up useless because
no examples reach them later on.  This prevents potentially useful
splits such as at leaf $j$. This problem can be solved by
recycling orphan nodes (subroutine \textbf{Swap} in
Algorithm~\ref{alg:OTT}). The general idea behind node recycling is to
allow nodes to split if a certain condition is met. In particular,
node $j$ splits if the following holds:
\vspace{-0.02in}
\begin{equation}
C_j - \max_{i \in \{1,2,\dots,k\}}{\bm l}_j(i) > R_S(C_r+1),
\label{eq:swap_cond}
\vspace{-0.02in}
\end{equation}
where $r$ denotes the root of the entire tree, $C_j$ is the size of the smallest leaf in the subtree with root $j$, where the smallest leaf is the one with the smallest total number of data points reaching it in the past, ${\bm l}_j$ is a $k$-dimensional vector of non-negative integers where the $i^{\text{th}}$ element is the count of the number of data points with label $i$ reaching leaf $j$ in the past, and finally $R_S$ is a ``swap resistance''. The subtraction of $\max_{i \in \{1,2,\dots,k\}}{\bm l}_j(i)$ in Equation~\ref{eq:swap_cond} ensures that a pure node will not be recycled. 

If the condition in Inequality~\ref{eq:swap_cond} is satisfied, the swap of the nodes is performed where an orphan leaf $s$, which was reached by the smallest number of examples in the past, and its parent $s_{\textsc{PA}}$ are detached from the tree and become children of node $j$ whereas the old sibling $s_{\textsc{sib}}$ of an orphan node $s$ becomes a direct child of the old grandparent $s_{\textsc{GPA}}$. The swapping procedure is shown in Figure~2. The condition captured in the Inequality~\ref{eq:swap_cond} allows us to prove that the number of times any given node is recycled is upper-bounded by the logarithm of the number of examples whenever the swap resistance is $4$ or more (Lemma~\ref{lemma:sb}).

\begin{lemma}
Let the swap resistance $R_S$ be greater or equal to $4$. Then for all sequences of examples, the number of times Algorithm~\ref{alg:OTT} recycles any given node is upper-bounded by the logarithm (with base $2$) of the sequence length.
\label{lemma:sb}
\end{lemma}

\section{Experiments}
\label{sec:experiments}
We address several hypotheses experimentally.
\vspace{-0.05in}
\begin{enumerate}
\item The LOMtree algorithm achieves true logarithmic time computation in practice.  
\vspace{-0.05in}
\item The LOMtree algorithm is competitive with or better than all other logarithmic train/test time algorithms for multiclass classification.
\vspace{-0.05in}
\item The LOMtree algorithm has statistical performance close to more common $O(k)$ approaches.
\end{enumerate}
\vspace{-0.2in}
\begin{wraptable}{l}{0.6\textwidth} 
\center
\vspace{-0.27in}
\setlength{\tabcolsep}{0.1pt}
\caption{Dataset sizes.}
\begin{tabular}{c|c|c|c|c|c|}
\cline{2-6}
& Isolet & Sector & Aloi & ImNet & ODP\\
  \hline
  \hline
\multicolumn{1}{|c||}{\multirow{1}{*}{size}} & 52.3\small{MB} & 19\small{MB}  & 17.7\small{MB} & 104\small{GB}\footnotemark[12] & 3\small{GB}\\
\hline
\multicolumn{1}{|c||}{\multirow{1}{*}{$\#$ features}} & 617 & 54\small{K} & 128 & 6144 & 0.5\small{M}\\
\hline
\multicolumn{1}{|c||}{\multirow{1}{*}{$\#$ examples}} &  7797 &  9619 &  108\small{K} & 14.2\small{M}& 1577418\\
\hline
\hline
\multicolumn{1}{|c||}{\multirow{1}{*}{$\#$ classes}} & 26 & 105 & 1000 & $\sim$22\small{K} & $\sim$105\small{K}\\
\hline
\end{tabular} 
\label{tab:dsize}
\vspace{-0.05in}
\end{wraptable}
\footnotetext[12]{compressed}

To address these hypotheses, we conducted experiments on a variety of benchmark multiclass datasets: \textit{Isolet}, \textit{Sector}, \textit{Aloi}, \textit{ImageNet} (\textit{ImNet}) and \textit{ODP}\footnote[13]{The details of the source of each dataset are provided in the Supplementary material.}. The details of the datasets are provided in Table~\ref{tab:dsize}. The datasets were divided into training ($90\%$) and testing ($10\%$). Furthermore, $10\%$ of the training dataset was used as a validation set. 

The baselines we compared \textit{LOMtree} with are a balanced random
tree of logarithmic depth (\textit{Rtree}) and the \textit{Filter
  tree}~\cite{BeygelzimerLR09}.  Where computationally feasible, we
also compared with a one-against-all classifier (\textit{OAA}) as a representative $O(k)$ approach.  All methods were implemented in the Vowpal Wabbit~\cite{VowpalWabbit} learning system and have similar levels of optimization. The regressors in the tree nodes
for \textit{LOMtree}, \textit{Rtree}, and \textit{Filter tree} as well as the \textit{OAA} regressors were trained by online gradient
descent for which we explored step sizes chosen from the set $\{0.25,0.5,0.75,1,2,4,8\}$. We used linear regressors. For each method we investigated training with up to $20$ passes through the data and we selected the best setting of the parameters (step size and number of passes) as the one minimizing the validation error. Additionally, for the \textit{LOMtree} we investigated different settings of the stopping criterion for the tree expansion: $T = \{k-1,2k-1,4k-1,8k-1,16k-1,32k-1,64k-1\}$, and swap resistance $R_S = \{4,8,16,32,64,128,256\}$.

In Table~\ref{tab:traintime} and~\ref{tab:peretesttime} we report respectively train time and per-example test time (the best performer is indicated in bold). Training time (and later reported test error) is not provided for \textit{OAA} on \textit{ImageNet} and \textit{ODP} due to intractability\footnote[14]{Note however that the mechanics of testing datastes are much easier - one can simply test with effectively untrained parameters on a few examples to measure the test speed thus the per-example test time for \textit{OAA} on \textit{ImageNet} and \textit{ODP} is provided.}-both are petabyte scale computations\footnote[15]{Also to the best of our knowledge there exist no state-of-the-art results of the \textit{OAA} performance on these datasets published in the literature.}. 
\vspace{-0.21in}
\begin{table}[h]
\hspace{-0.12in}\begin{minipage}[t]{0.5\textwidth}
\center
\setlength{\tabcolsep}{4pt}
\caption{Training time on selected problems.}
\begin{tabular}{c|c|c|c|}
  \cline{2-4} 
\multirow{2}{*}{} & Isolet & Sector & Aloi\\
  \hline
  \hline
\multicolumn{1}{|c||}{\multirow{1}{*}{LOMtree}} & \textbf{16.27s} & \textbf{12.77s} & \textbf{51.86s}\\
  \hline
\multicolumn{1}{|c||}{\multirow{1}{*}{OAA}} & 19.58s & 18.37s & 11m2.43s\\
\hline
\end{tabular} 
\label{tab:traintime}
\end{minipage}
\hspace{-0.12in}\begin{minipage}[t]{0.5\textwidth}
\center
\setlength{\tabcolsep}{1.6pt}
\caption{Per-example test time on all problems.}
\vspace{-0.1in}
\begin{tabular}{c|c|c|c|c|c|}
  \cline{2-6} 
\multirow{2}{*}{} & Isolet & Sector & Aloi & ImNet & ODP\\
  \hline
  \hline
\multicolumn{1}{|c||}{\multirow{1}{*}{LOMtree}} & \textbf{0.14ms} & \textbf{0.13ms} & \textbf{0.06ms} & \textbf{0.52ms} & \textbf{0.26ms}\\
  \hline
\multicolumn{1}{|c||}{\multirow{1}{*}{OAA}} & 0.16 ms & 0.24ms & 0.33ms & 0.21s & 1.05s\\
\hline
\end{tabular} 
\label{tab:peretesttime}
\end{minipage}
\vspace{-0.19in}
\end{table}

\begin{wrapfigure}{l}{0.5\textwidth} 
  \begin{center}
\includegraphics[scale = 0.3]{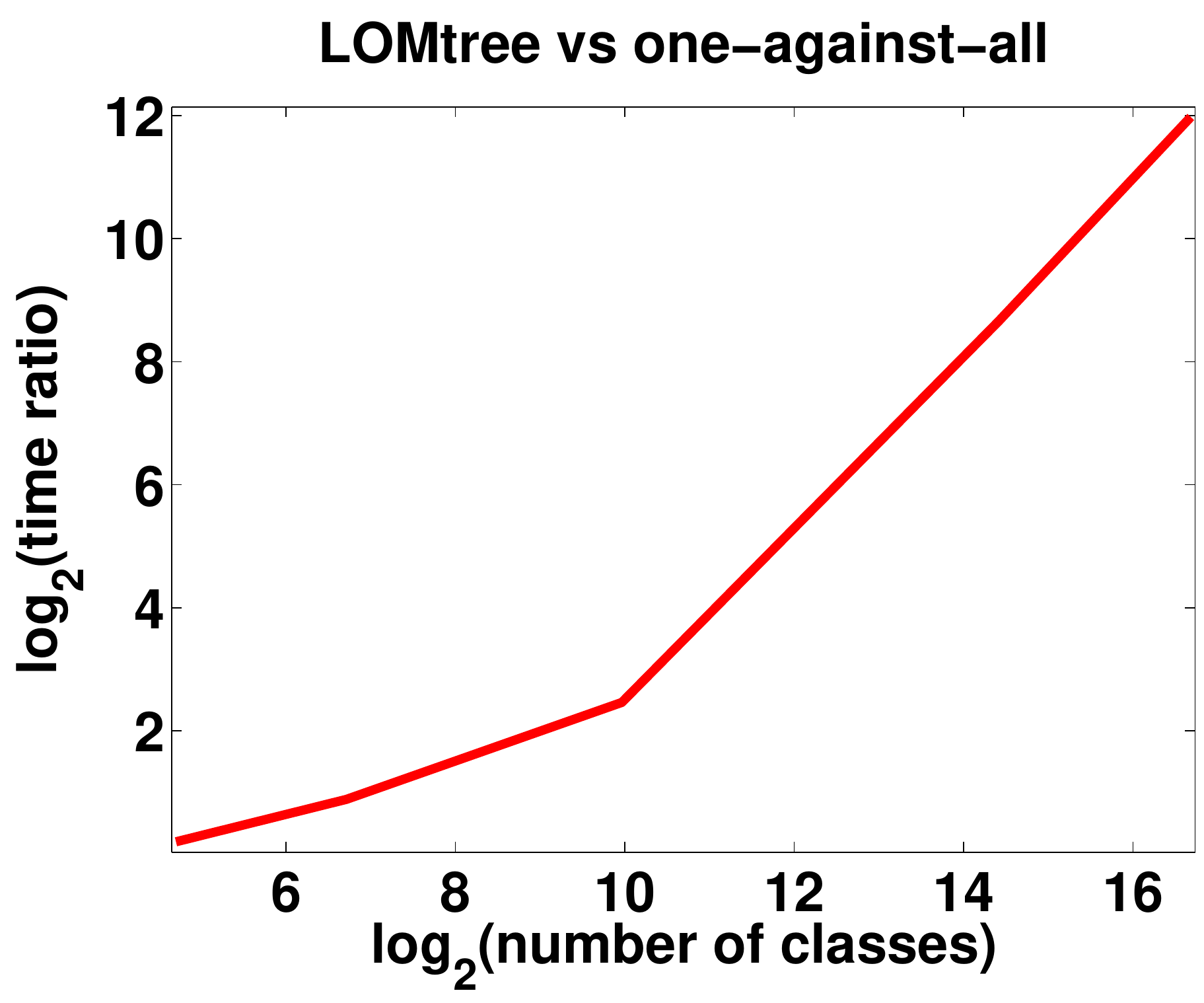}
\end{center}
\vspace{-0.24in}
\caption{Logarithm of the ratio of per-example test times of \textit{OAA} and \textit{LOMtree} on all problems.}
\label{fig:impex}
\vspace{-0.4in}
\end{wrapfigure}

The first hypothesis is consistent with the experimental results.  Time-wise \textit{LOMtree} significantly outperforms \textit{OAA} due to building only close-to logarithmic depth trees.  The improvement in the training time increases with the number of classes in the classification problem.  For instance on \textit{Aloi} training with \textit{LOMtree} is $12.8$ times faster than with $OAA$. The same can be said about the test time, where the per-example test time for \textit{Aloi}, \textit{ImageNet} and \textit{ODP} are respectively $5.5$, $403.8$ and $4038.5$ times faster than \textit{OAA}. The significant advantage of \textit{LOMtree} over \textit{OAA} is also captured in Figure~\ref{fig:impex}.
\vspace{-0.16in}

Next, in Table~\ref{tab:testerr} (the best logarithmic time performer is indicated in bold) we report test error of logarithmic train/test time algorithms. We also show the binomial symmetrical $95\%$ confidence intervals for our results. Clearly the second hypothesis is also consistent with the experimental results.  Since the \textit{Rtree} imposes a random label partition, the resulting error it obtains is generally worse than the error obtained by the competitor methods including \textit{LOMtree} which learns the label partitioning directly from the data. At the same time \textit{LOMtree} beats \textit{Filter tree} on every dataset, though for \textit{ImageNet} and \textit{ODP} (both have a high level of noise) the advantage of \textit{LOMtree} is not as significant. 

\vspace{-0.12in}
\begin{table}[htp!]
\center
\setlength{\tabcolsep}{4pt}
\vspace{-0.05in}
\caption{Test error ($\%$) and confidence interval on all problems.}
\begin{tabular}{c|c|c|c||c|}
  \cline{2-5} 
\multirow{2}{*}{} & LOMtree & Rtree & Filter tree & OAA\\
  \hline
  \hline
\multicolumn{1}{|c||}{\multirow{1}{*}{Isolet}} & \textbf{6.36}\small{$\pm{1.71}$} & 16.92\small{$\pm{2.63}$} & 15.10\small{$\pm{2.51}$} & $3.56$\small{$\pm 1.30\%$}\\
  \hline
\multicolumn{1}{|c||}{\multirow{1}{*}{Sector}} & 16.19\small{$\pm{2.33}$} & \textbf{15.77}\small{$\pm{2.30}$} & 17.70\small{$\pm{2.41}$} & $9.17$\small{$\pm{1.82\%}$}\\
 \hline
\multicolumn{1}{|c||}{\multirow{1}{*}{Aloi}} & \textbf{16.50}\small{$\pm{0.70}$} & 83.74\small{$\pm{0.70}$} & 80.50\small{$\pm{0.75}$} & $13.78$\small{$\pm{0.65\%}$}\\
 \hline
\multicolumn{1}{|c||}{\multirow{1}{*}{ImNet}} & \textbf{90.17}\small{$\pm{0.05}$} & 96.99\small{$\pm{0.03}$} & 92.12\small{$\pm{0.04}$} & NA\\
 \hline
\multicolumn{1}{|c||}{\multirow{1}{*}{ODP}} & \textbf{93.46}\small{$\pm{0.12}$} & 93.85\small{$\pm{0.12}$} & 93.76\small{$\pm{0.12}$} & NA\\
\hline
\end{tabular} 
\label{tab:testerr}
\vspace{-0.05in}
\end{table}

The third hypothesis is weakly consistent with the empirical results. The time advantage of \textit{LOMtree} comes with some loss of statistical accuracy with respect to \textit{OAA} where \textit{OAA} is tractable. We conclude that \textit{LOMtree} significantly closes the gap between other logarithmic time methods and \textit{OAA}, making it a plausible approach in computationally constrained large-$k$ applications.

\vspace{-0.11in}
\section{Conclusion}
The LOMtree algorithm reduces the multiclass problem to a set of
binary problems organized in a tree structure where the partition
in every tree node is done by optimizing a new partition criterion online. The criterion guarantees pure and balanced splits
leading to logarithmic training and testing time for the tree
classifier. We provide theoretical justification for our approach via a boosting
statement and empirically evaluate it on multiple multiclass
datasets.  Empirically, we find that this is the best available
logarithmic time approach for multiclass classification problems.

\subsubsection*{Acknowledgments}

We would like to thank Alekh Agarwal, Dean Foster, Robert Schapire and Matus Telgarsky for valuable discussions.



\small{
\bibliographystyle{unsrt}
\bibliography{PAPER_LOM}

\begin{thebibliography}{10}

\bibitem{Rifkin2004}
R.~Rifkin and A.~Klautau.
\newblock In defense of one-vs-all classification.
\newblock {\em J. Mach. Learn. Res.}, 5:101--141, 2004.

\bibitem{CnT}
T.~M. Cover and J.~A. Thomas.
\newblock {\em Elements of Information Theory}.
\newblock John Wiley \& Sons, Inc., 1991.

\bibitem{ig}
L.~Breiman, J.~H. Friedman, R.~A. Olshen, and C.~J. Stone.
\newblock {\em Classification and Regression Trees}.
\newblock CRC Press LLC, Boca Raton, Florida, 1984.

\bibitem{Kearns95}
M.~Kearns and Y.~Mansour.
\newblock On the boosting ability of top-down decision tree learning
  algorithms.
\newblock {\em Journal of Computer and Systems Sciences}, 58(1):109--128, 1999
  (also In \textit{STOC}, 1996).

\bibitem{BeygelzimerLR09}
A.~Beygelzimer, J.~Langford, and P.~D. Ravikumar.
\newblock Error-correcting tournaments.
\newblock In {\em ALT}, 2009.

\bibitem{BeygelzimerLLSS09}
A.~Beygelzimer, J.~Langford, Y.~Lifshits, G.~B. Sorkin, and A.~L. Strehl.
\newblock Conditional probability tree estimation analysis and algorithms.
\newblock In {\em UAI}, 2009.

\bibitem{Bishop:2006:PRM:1162264}
C.~M. Bishop.
\newblock {\em Pattern Recognition and Machine Learning}.
\newblock Springer, 2006.

\bibitem{BengioWG10}
S.~Bengio, J.~Weston, and D.~Grangier.
\newblock Label embedding trees for large multi-class tasks.
\newblock In {\em NIPS}, 2010.

\bibitem{journals/informaticaSI/MadzarovGC09}
G.~Madzarov, D.~Gjorgjevikj, and I.~Chorbev.
\newblock A multi-class svm classifier utilizing binary decision tree.
\newblock {\em Informatica}, 33(2):225--233, 2009.

\bibitem{DengSBL11}
J.~Deng, S.~Satheesh, A.~C. Berg, and L.~Fei-Fei.
\newblock Fast and balanced: Efficient label tree learning for large scale
  object recognition.
\newblock In {\em NIPS}, 2011.

\bibitem{weston13}
J.~Weston, A.~Makadia, and H.~Yee.
\newblock Label partitioning for sublinear ranking.
\newblock In {\em ICML}, 2013.

\bibitem{conf/cvpr/ZhaoX13}
B.~Zhao and E.~P. Xing.
\newblock Sparse output coding for large-scale visual recognition.
\newblock In {\em CVPR}, 2013.

\bibitem{DBLP:journals/corr/abs-0902-1284}
D.~Hsu, S.~Kakade, J.~Langford, and T.~Zhang.
\newblock Multi-label prediction via compressed sensing.
\newblock In {\em NIPS}, 2009.

\bibitem{DBLP:journals/corr/AgarwalKKSV13}
A.~Agarwal, S.~M. Kakade, N.~Karampatziakis, L.~Song, and G.~Valiant.
\newblock Least squares revisited: Scalable approaches for multi-class
  prediction.
\newblock In {\em ICML}, 2014.

\bibitem{icml2014c2_beijbom14}
O.~Beijbom, M.~Saberian, D.~Kriegman, and N.~Vasconcelos.
\newblock Guess-averse loss functions for cost-sensitive multiclass boosting.
\newblock In {\em ICML}, 2014.

\bibitem{Manik}
R.~Agarwal, A.~Gupta, Y.~Prabhu, and M.~Varma.
\newblock Multi-label learning with millions of labels: Recommending advertiser
  bid phrases for web pages.
\newblock In {\em WWW}, 2013.

\bibitem{Prabhu2014}
Y.~Prabhu and M.~Varma.
\newblock Fastxml: A fast, accurate and stable tree-classifier for extreme
  multi-label learning.
\newblock In {\em ACM SIGKDD}, 2014.

\bibitem{LMLML14}
H.-F. Yu, P.~Jain, P.~Kar, and I.~S. Dhillon.
\newblock Large-scale multi-label learning with missing labels.
\newblock In {\em ICML}, 2014.

\bibitem{LiuLargeScale2005}
T.-Y. Liu, Y.~Yang, H.~Wan, H.-J. Zeng, Z.~Chen, and W.-Y. Ma.
\newblock Support vector machines classification with a very large-scale
  taxonomy.
\newblock In {\em SIGKDD Explorations}, 2005.

\bibitem{conf/sigir/BennettN09}
P.~N. Bennett and N.~Nguyen.
\newblock Refined experts: improving classification in large taxonomies.
\newblock In {\em SIGIR}, 2009.

\bibitem{MTSWIMC}
A.~Montillo, J.~Tu, J.~Shotton, J.~Winn, J.E. Iglesias, D.N. Metaxas, and
  A.~Criminisi.
\newblock Entanglement and differentiable information gain maximization.
\newblock {\em Decision Forests for Computer Vision and Medical Image
  Analysis}, 2013.

\bibitem{Tentori2007107}
K.~Tentori, V.~Crupi, N.~Bonini, and D.~Osherson.
\newblock Comparison of confirmation measures.
\newblock {\em Cognition}, 103(1):107 -- 119, 2007.

\bibitem{Carnap1962}
R.~Carnap.
\newblock {\em Logical Foundations of Probability. 2nd ed.}
\newblock Chicago: University of Chicago Press. Par. 87 (pp. 468-478), 1962.

\bibitem{ShaiSS2012}
S.~Shalev-Shwartz.
\newblock Online learning and online convex optimization.
\newblock {\em Found. Trends Mach. Learn.}, 4(2):107--194, 2012.

\bibitem{VowpalWabbit}
J.~Langford, L.~Li, and A.~Strehl.
\newblock \url{ http://hunch.net/~vw}, 2007.

\bibitem{Nesterov2004}
Y.~Nesterov.
\newblock Introductory lectures on convex optimization : a basic course.
\newblock {\em Applied optimization, Kluwer Academic Publ.}, 2004.

\bibitem{Deng09imagenet:a}
J.~Deng, W.~Dong, R.~Socher, L.-J. Li, K.~Li, and L.~Fei-Fei.
\newblock Imagenet: A large-scale hierarchical image database.
\newblock In {\em CVPR}, 2009.

\end{thebibliography}
}

\newpage
\normalsize

\vbox{\hsize\textwidth
\linewidth\hsize \vskip 0.1in \toptitlebar \centering
{\LARGE\bf Logarithmic Time Online Multiclass prediction\\(Supplementary Material) \par}
\bottomtitlebar
\vskip 0.3in minus 0.1in}
\vspace{-0.3in}

\section{Bottom-up partitions do not work}
\label{sec:bottom-up}
The most natural bottom-up construction for creating partitions is not
viable as will be now shown by an example. Bottom-up construction techniques start by pairing labels, either
randomly or arbitrarily, and then building a predictor of whether the
class label is left or right conditioned on the class label being one
of the paired labels.  In order to construct a full tree, this
operation must compose, pairing trees with size $2$ to create trees of
size $4$. Here, we show that the straightforward approach to
composition fails.

Suppose we have a one dimensional feature space with examples of class
label $i$ having feature value $i$ and we work with threshold
predictors.  Suppose we have 4 classes $1, 2, 3, 4$, and we happen to
pair $(1,3)$ and $(2,4)$.  It is easy to build a linear predictor for
each of these splits.  The next step is building a predictor for
$(1,3)$ vs $(2,4)$ which is impossible because all thresholds in
$(-\infty,1)$, $(2,3)$, and $(4,\infty)$ err on two labels while
thresholds on $(1,2)$ and $(3,4)$ err on one label.

\section{Proof of Lemma~\ref{lemma:obj-to-purity}}

We start from deriving an upper-bound on $J(h)$. For the ease of notation let $P_i = P(h(x) > 0 | i)$. Thus
\[J(h) = 2\sum_{i=1}^{k}\pi_i\left \lvert P(h(x) > 0 | i) - P(h(x) > 0)\right \rvert = 2\sum_{i=1}^{k}\pi_i\left \lvert P_i - \sum_{j=1}^{k}\pi_jP_j\right \rvert,
\]
where $\forall_{i = \{1,2,\dots,k\}}0 \leq P_i \leq 1$. Let $\alpha_i = \min(P_i,1-P_i)$ and recall the purity factor $\alpha = \sum_{i=1}^{k}\pi_i\alpha_i$ and the balancing factor $\beta = P(h(x) > 0)$. Without loss of generality let $\beta \leq \frac{1}{2}$. Furthermore, let
\[L_1 = \{i:i \in \{1,2,\dots,k\}, P_i \geq \frac{1}{2}\}, \:\:\:L_2 = \{i:i \in \{1,2,\dots,k\}, P_i \in [\beta,\frac{1}{2})\}
\]
\[\text{and} \:\:\:\:\:L_3 = \{i:i \in \{1,2,\dots,k\}, P_i < \beta\}.
\]
First notice that
\begin{equation}
\beta = \sum_{i=1}^{k}\pi_iP_i = \sum_{i \in L_1}\pi_i(1 - \alpha_i) + \sum_{i \in L_2 \cup L_3}\pi_i\alpha_i = \sum_{i \in L_1}\pi_i - 2\sum_{i \in L_1}\pi_i\alpha_i + \alpha
\label{eq:medium}
\end{equation}
Therefore 
\begin{eqnarray*}
\frac{J(h)}{2} &=& \sum_{i=1}^{k}\pi_i\left \lvert P_i - \beta \right \rvert = \sum_{i \in L_1}\pi_i(1 - \alpha_i - \beta) + \sum_{i \in L_2}\pi_i(\alpha_i - \beta) + \sum_{i \in L_3}\pi_i(\beta - \alpha_i)\\
&=& \sum_{i \in L_1}\pi_i(1 - \beta) - \sum_{i \in L_1}\pi_i\alpha_i + \sum_{i \in L_2}\pi_i\alpha_i - \sum_{i \in L_2}\pi_i\beta + \sum_{i \in L_3}\pi_i\beta - \sum_{i \in L_3}\pi_i\alpha_i
\end{eqnarray*}
Note that $\sum_{i \in L_3}\pi_i = 1 - \sum_{i \in L_1}\pi_i - \sum_{i \in L_2}\pi_i$ and therefore
\begin{eqnarray*}
\frac{J(h)}{2} &=& \sum_{i \in L_1}\pi_i(1 \!-\! \beta) \!-\!\!\! \sum_{i \in L_1}\pi_i\alpha_i \!+\!\!\! \sum_{i \in L_2}\pi_i\alpha_i \!-\!\!\! \sum_{i \in L_2}\pi_i\beta + \beta(1 \!-\!\!\! \sum_{i \in L_1}\pi_i \!-\!\!\! \sum_{i \in L_2}\pi_i) \!-\!\!\! \sum_{i \in L_3}\pi_i\alpha_i\\
&=& \sum_{i \in L_1}\pi_i(1 - 2\beta) - \sum_{i \in L_1}\pi_i\alpha_i + \sum_{i \in L_2}\pi_i\alpha_i + \beta(1 - 2\sum_{i \in L_2}\pi_i) - \sum_{i \in L_3}\pi_i\alpha_i
\end{eqnarray*}
Furthermore, since $- \sum_{i \in L_1}\pi_i\alpha_i + \sum_{i \in L_2}\pi_i\alpha_i - \sum_{i \in L_3}\pi_i\alpha_i = - \alpha + 2\sum_{i \in L_2}\pi_i\alpha_i$ we further write that
\begin{eqnarray*}
\frac{J(h)}{2} &=& \sum_{i \in L_1}\pi_i(1 - 2\beta)+ \beta(1 - 2\sum_{i \in L_2}\pi_i) - \alpha + 2\sum_{i \in L_2}\pi_i\alpha_i
\end{eqnarray*}
By Equation~\ref{eq:medium}, it can be further rewritten as
\begin{eqnarray*}
\frac{J(h)}{2} &=& (1 - 2\beta)(\beta + 2\sum_{i \in L_1}\pi_i\alpha_i - \alpha)+ \beta(1 - 2\sum_{i \in L_2}\pi_i) - \alpha + 2\sum_{i \in L_2}\pi_i\alpha_i\\
&=& 2(1 - \beta)(\beta - \alpha) + 2(1 - 2\beta)\sum_{i \in L_1}\pi_i\alpha_i + 2\sum_{i \in L_2}\pi_i(\alpha_i - \beta)
\end{eqnarray*}
Since $\alpha_i$'s are bounded by $0.5$ we obtain
\begin{eqnarray*}
\frac{J(h)}{2} &\leq& 2(1 - \beta)(\beta - \alpha) + 2(1 - 2\beta)\sum_{i \in L_1}\pi_i\alpha_i + 2\sum_{i \in L_2}\pi_i(\frac{1}{2} - \beta)\\
&\leq& 2(1 - \beta)(\beta - \alpha) + 2(1 - 2\beta)\alpha + 1 - 2\beta\\
&=& 2\beta(1 - \beta) - 2\alpha(1 - \beta) + 2\alpha(1 - 2\beta) + 1 - 2\beta\\
&=& 1 - 2\beta^2 - 2\beta\alpha
\end{eqnarray*}
Thus:
\[\alpha \leq \frac{2 - J(h)}{4\beta} - \beta.
\]

\section{Proof of Lemma~\ref{lemma:maximal}}

\begin{proof}
We first show that $J(h) \in [0,1]$. We start from deriving an upper-bound on $J(h)$, where $h \in \mathcal{H}$ is some hypothesis in the hypothesis class. For the ease of notation let $P_i = P(h(x) > 0 | i)$. Thus
\begin{eqnarray}
J(h) &=& 2\sum_{i=1}^{k}\pi_i\left \lvert P(h(x) > 0 | i) - P(h(x) > 0)\right \rvert\\
& = &2\sum_{i=1}^{k}\pi_i\left \lvert P_i - \sum_{j=1}^{k}\pi_jP_j\right \rvert, \nonumber
\label{eq:objform}
\end{eqnarray}
where $\forall_{i = \{1,2,\dots,k\}}0 \leq P_i \leq 1$. The objective $J(h)$ is certainly maximized on the extremes of the $[0,1]$ interval. The upper-bound on $J(h)$ can be thus obtained by setting some of the $P_i$'s to $1$'s and remaining ones to $0$'s. To be more precise, let 
\[L_1 = \{i:i \in \{1,2,\dots,k\}, P_i = 1\} \text{\:\:\:\:\:\:and\:\:\:\:\:\:} L_2 = \{i:i \in \{1,2,\dots,k\}, P_i = 0\}.
\]
Therefore it follows that
\begin{eqnarray*}
J(h) &\leq& 2\left[\sum_{i \in L_1} \pi_i(1 - \sum_{j \in L_1}\pi_j) + \sum_{i \in L_2} \pi_i\sum_{j \in L_1}\pi_j\right]\\
&=&  2\left[\sum_{i \in L_1} \pi_i - ( \sum_{i \in L_1} \pi_i)^2 + (1 -  \sum_{i \in L_1} \pi_i) \sum_{i \in L_1} \pi_i\right]\\ 
&=& 4\left[\sum_{i \in L_1} \pi_i - ( \sum_{i \in L_1} \pi_i)^2\right]
\end{eqnarray*}

Let $b = \sum_{i \in L_1} \pi_i$ thus 
\begin{equation}
J(h) \leq 4b(1 - b) = -4b^2 + 4b
\label{eq:upper_bound}
\end{equation} 

Since $b \in [0,1]$, it is straightforward that $-4b^2 + 4b \in [0,1]$ and thus $J(h) \in [0,1]$. 

We now proceed to prove the main statement of Lemma~\ref{lemma:maximal}, if $h$ induces a maximally pure and balanced partition then $J(h) = 1$. Since $h$ is maximally balanced, $P(h(x) > 0) = 0.5$. Simultaneously, since $h$ is maximally pure $\forall_{i = \{1,2,\dots,k\}}(P(h(x) > 0|i) = 0 \:\:\text{or}\:\: P(h(x) > 0|i) = 1)$. Substituting that into Equation~\ref{eq:objform} yields that $J(h) = 1$.
\end{proof}

\section{Proof of Theorem~\ref{thm:main}}
\label{sec:maindetails}

\begin{proof}

The analysis studies a tree construction algorithm where we recursively find the leaf node with the highest weight, and choose to
split it into two children. Consider the tree constructed over $t$
steps where in each step we take one leaf node and split it into two. Let $n$ be the
heaviest node at time $t$ and its weight $w_n$ be denoted by $w$ for
brevity. Consider splitting this leaf to two children $n_0$ and
$n_1$. For the ease of notation let $w_0 = w_{n_0}$ and $w_1 =
w_{n_1}$. Also for the ease of notation let $\beta = P(h_n(x) > 0)$
and $P_i = P(h_n(x) > 0|i)$. Let $\pi_i$ be the shorthand for
$\pi_{n,i}$ and $h$ be the shorthand for $h_n$. Recall that $\beta = \sum_{i=1}^k\pi_iP_i$ and
$\sum_{i=1}^k\pi_i = 1$. Also notice that $w_0 = w(1-\beta)$ and $w_1
= w\beta$. Let ${\bm \pi}$ be the $k$-element vector with $i^{th}$
entry equal to $\pi_i$. Furthermore let $\tilde{G}({\bm \pi}) = \sum_{i =
  1}^k \pi_{i}\ln \left( \frac{1}{\pi_{i}} \right)$. 

Before the split the contribution of node $n$ to
$G_t$ was $w\tilde{G}({\bm \pi})$.  Let $\pi_{n_0,i} =
\frac{\pi_i(1 - P_i)}{1 - \beta}$ and $\pi_{n_1,i} =
\frac{\pi_iP_i}{\beta}$ be the probabilities that a randomly chosen $x$
drawn from $\mathcal{P}$ has label $i$ given that $x$ reaches nodes
$n_0$ and $n_1$ respectively. For brevity, let $\pi_{n_0,i}$ be denoted by $\pi_{0,i}$ and $\pi_{n_1,i}$ be denoted by $\pi_{1,i}$. Furthermore let ${\bm \pi}_0$ be the
$k$-element vector with $i^{th}$ entry equal to $\pi_{0,i}$ and let
${\bm \pi}_1$ be the $k$-element vector with $i^{th}$ entry equal
to $\pi_{1,i}$. Notice that ${\bm \pi} = (1 - \beta){\bm \pi}_0 +
\beta{\bm \pi}_1$. After the split the contribution of the same,
now internal, node $n$ changes to $w((1- \beta)\tilde{G}({\bm \pi}_0) + \beta
\tilde{G}({\bm \pi}_1))$. We denote the difference between them as
$\Delta_t$ and thus
\vspace{-0.02in}
\begin{equation}
  \Delta_t := G_t - G_{t+1} = w\left[\tilde{G}({\bm \pi}) - (1- \beta)\tilde{G}({\bm
    \pi}_0) - \beta \tilde{G}({\bm \pi}_1)\right].
  \label{eqn:ent-decrease}
\vspace{-0.02in}
\end{equation}
We aim to lower-bound $\Delta_t$. The entropy reduction of
Equation~\ref{eqn:ent-decrease}~\cite{Kearns95} corresponds to a gap
in the Jensen's inequality applied to the concave function $\tilde{G}(\bm
\pi)$. This leads to the lower-bound on $\Delta_t$ given in Lemma~\ref{lem:lower-bound} (the lemma is proven in Section~\ref{sec:lower-boundproof} in the Supplementary material).
\begin{lemma}
The entropy reduction $\Delta_t$ of Equation~\ref{eqn:ent-decrease} can be lower-bounded as follows
\[\Delta_t \geq
\frac{J(h)^2G_t}{8\beta(1-\beta)t\ln k}
\]
\label{lem:lower-bound}
\vspace{-0.15in}
\end{lemma}

Lemma~\ref{lem:lower-bound} implies that the larger the objective
$J(h)$ is at time $t$, the larger the entropy reduction ends up being,
which further reinforces intuitions to maximize $J$. In general, it
might not be possible to find any hypothesis with a large enough
objective $J(h)$ to guarantee sufficient progress at this point so we
appeal to a \emph{weak learning assumption}. This assumption can be used to further lower-bound $\Delta_t$. The lower-bound can then be used (details are in Section~\ref{sec:maindetails} in the Supplementary material) to obtain the main theoretical statement of the paper captured in Theorem~\ref{thm:main}.

From the definition of $\gamma$ it follows that $1 - \gamma \geq \beta \geq \gamma$. Also note that the
\emph{weak hypothesis assumption} guarantees $J(h) \geq 2\gamma$, which applied to the lower-bound on $\Delta_t$ captured in Lemma~\ref{lem:lower-bound} yields
\vspace{-0.05in}
\[\Delta_t \geq \frac{\gamma^2G_t}{2(1 - \gamma)^2t\ln k}.
\vspace{-0.05in}
\]
Let $\eta = \sqrt{\frac{8}{(1 - \gamma)^2\ln k}}\gamma$. Then $\Delta_t > \frac{\eta^2G_t}{16t}$. Thus we obtain the recurrence inequality
\vspace{-0.05in}
\[G_{t+1} \leq G_t - \Delta_t < G_t - \frac{\eta^2G_t}{16t} =
G_t\left[1 - \frac{\eta^2}{16t}\right] 
\vspace{-0.02in}
\]
One can now compute the minimum number of splits required to reduce
$G_t$ below $\alpha$, where $\alpha \in [0,1]$. Applying the proof technique from~\cite{Kearns95} (the proof of Theorem 10) gives the final statement of Theorem~\ref{thm:main}.
\end{proof}

\section{Proof of Lemma~\ref{lem:lower-bound}}
\label{sec:lower-boundproof}

\begin{proof}
Without loss of generality
assume that $P_1 \leq P_2 \leq \dots \leq P_k$. As mentioned before, the entropy reduction $\Delta_t$ corresponds to a gap in the Jensen's
inequality applied to the concave function $\tilde{G}(\bm \pi)$. Also recall that Shannon entropy is strongly concave with respect to $\ell_1$-norm (see e.g., Example 2.5 in
Shalev-Shwartz~\cite{ShaiSS2012}). As a specific consequence (see
e.g. Theorem 2.1.9 in Nesterov~\cite{Nesterov2004}) we obtain
\begin{equation}
\Delta_t \geq w\beta(1-\beta)\|{\bm \pi}_0 - {\bm \pi}_1\|_1^2 = \frac{w}{\beta(1-\beta)}\left(\sum_{i
  = 1}^k\left|\pi_i(P_i - \beta)\right|\right)^2 =
\frac{wJ(h)^2}{4\beta(1 - \beta)},
\label{eq:subst}
\end{equation}
where the last equality results from the definition of $J(h) = 2\sum_{i =
  1}^k\pi_i|P_i - \beta|$.

Note that the following holds $w \geq
\frac{G_t}{2t\ln k}$, where recall that $w$ is the weight of the heaviest leaf in the tree, i.e. the leaf with the highest weight, at round $t$. This leaf is selected to the currently considered split~\cite{Kearns95}. In particular, the lower-bound on $w$ is the consequence of the following
\[ G_t \!=\! \sum_{l \in \mathcal{L}}\!\!w_l\!\sum_{i = 1}^k\!\! \pi_{l,i}\ln \left( \frac{1}{\pi_{l,i}} \right)
\leq \sum_{l \in \mathcal{L}}\!\!w_l\ln k \leq 2tw\ln k,
\]
where $w = \max_{l \in \mathcal{L}}w_l$. Thus $w \geq \frac{G_t}{2t\ln
  k}$ which when substituted to Equation~\ref{eq:subst} gives the final statement of the lemma.
\end{proof}

\section{Proof of Lemma~\ref{lemma:sb}}

\begin{proof}
We bound the number of swaps that any node makes. Consider $R_S =
4$ and let $j$ be the node that is about to split and $s$ be
the orphan node that will be recycled (thus $C_r = C_s$). The
condition in Equation~\ref{eq:swap_cond} implies that the swap is done
if $C_j > 4(C_r + 1) = 4(C_s + 1)$. Algorithm~\ref{alg:OTT} makes $s$
a child of $j$ during the swap and sets its counter to $C_s^{new} =
\floor*{C_j/2} \geq 2(C_r + 1) = 2(C_s + 1)$. Then $C_r$ gets updated. Since the value of $C_s^{new}$ at least doubles after a
swap and all counters are bounded by the number of examples $n$, the node
can be involved in at most $\log_2 n$ swaps.
\end{proof}

\section{Equivalent forms of the objective function}
\label{sec:explanation}

Consider the objective function as given in Equation~\ref{eqn:objective}
\[J(h) = 2\sum_{i=1}^k \pi_i \left| P(h(x) > 0) - P(h(x) > 0 | i)
  \right|.
\]
Recall that $\mathcal{X}$ denotes the set of all examples and let $\mathcal{X}_i$ denote the set of examples in class $i$. Also let $|\mathcal{X}|$ denote the cardinality of set $\mathcal{X}$ and let $|\mathcal{X}_i|$ denote the cardinality of set $\mathcal{X}_i$. Then we can re-write the objective as
\begin{eqnarray}
J(h) &=& 2\sum_{i=1}^k \pi_i \left| \frac{\sum_{x \in \mathcal{X}}\mathds{1}(h(x) > 0)}{|\mathcal{X}|} - \frac{\sum_{x \in \mathcal{X}_i}\mathds{1}(h(x) > 0)}{|\mathcal{X}_i|}
  \right| \nonumber\\
&=& 2\sum_{i=1}^k \pi_i \left| \mathbb{E}_x[\mathds{1}(h(x) > 0)] - \mathbb{E}_{x}[\mathds{1}(h(x) > 0|i)]
  \right| \nonumber\\
&=& 2\mathbb{E}_i[\left| \mathbb{E}_x[\mathds{1}(h(x) > 0)] - \mathbb{E}_{x}[\mathds{1}(h(x) > 0|i)]
  \right|]. \nonumber
\end{eqnarray}

\newpage

\section{Toy example of the behavior of LOMtree algorithm}
\label{sec:toy}

\begin{figure}[h]
\center
a) \includegraphics[width = 1.6in]{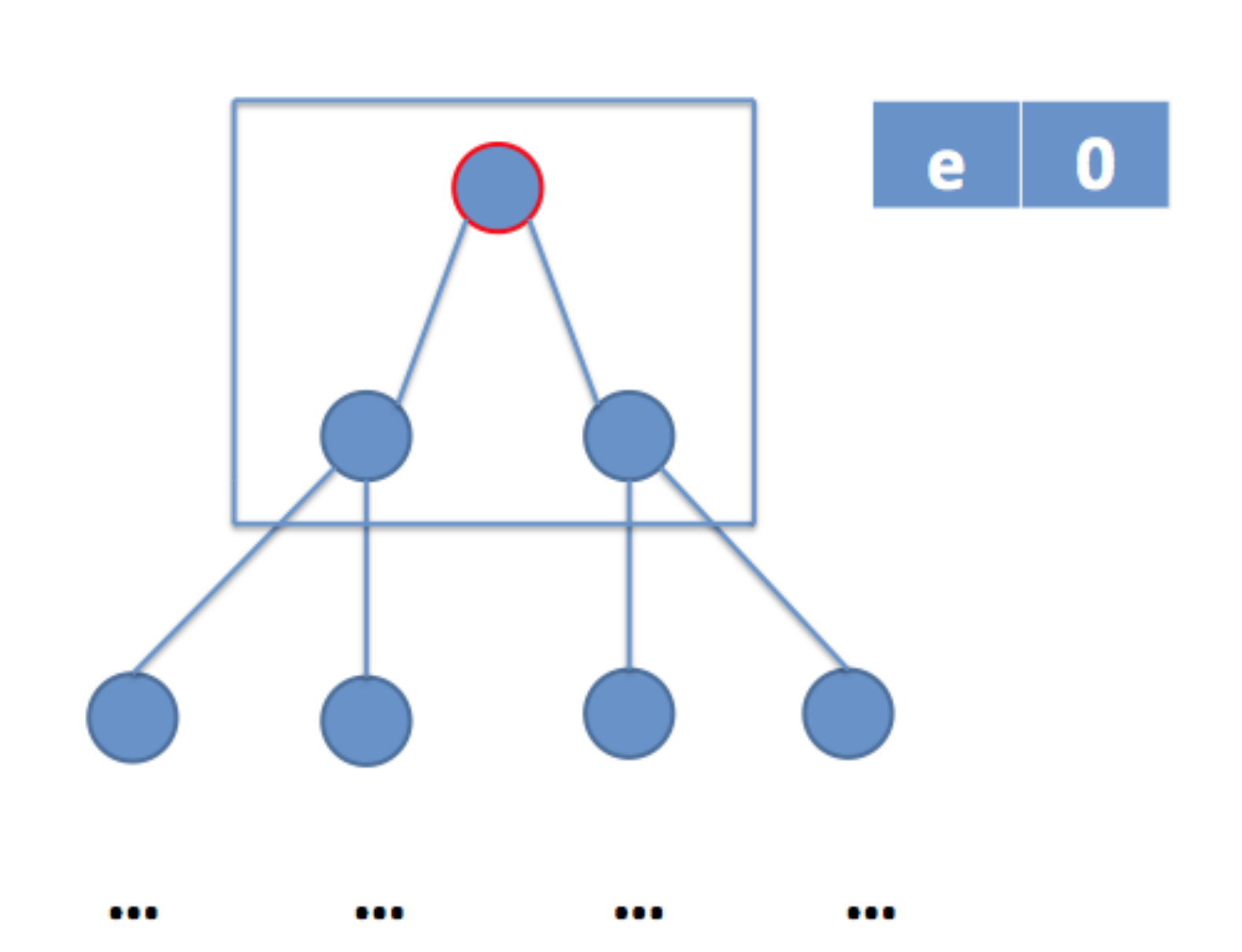}
b) \includegraphics[width = 1.6in]{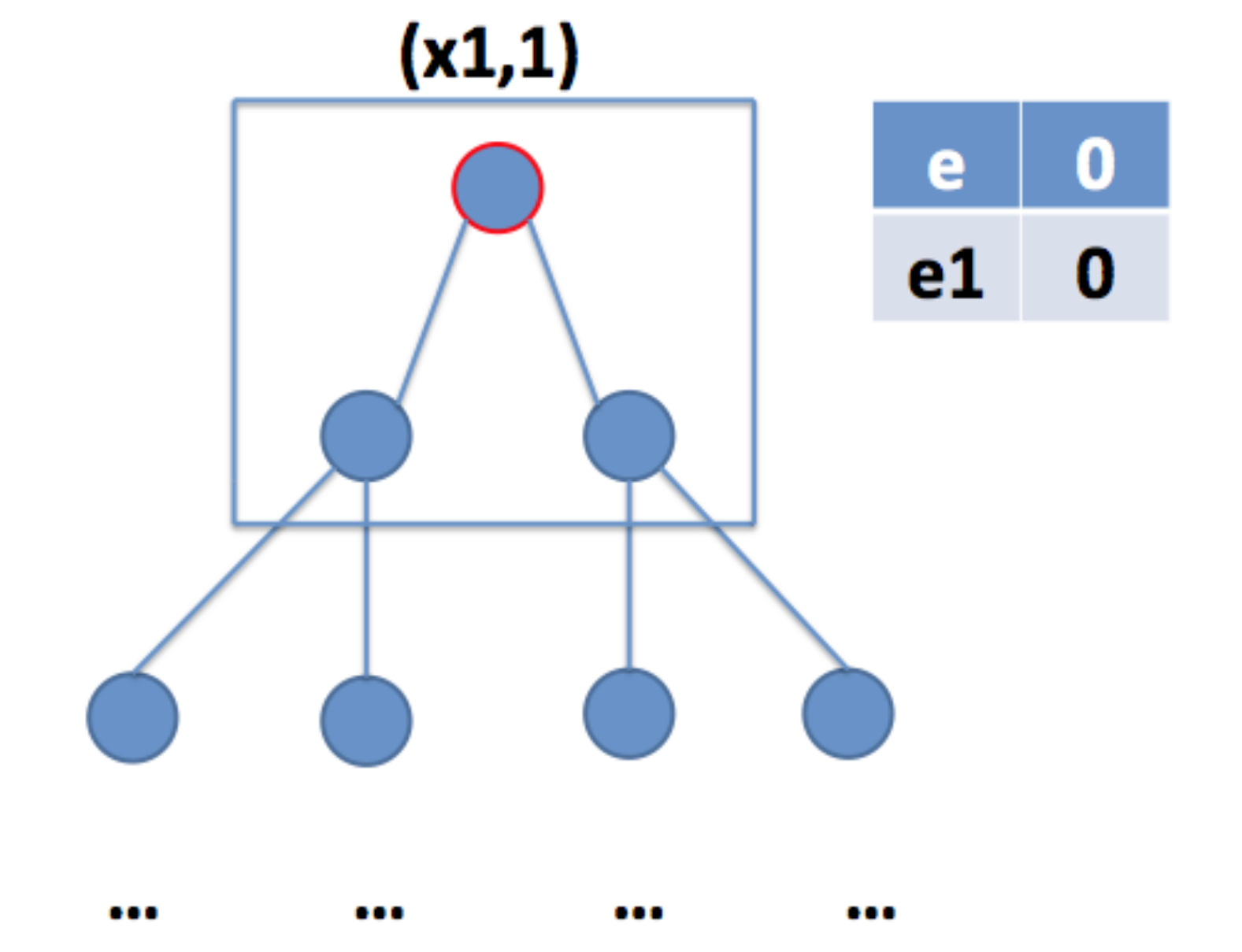}
c) \includegraphics[width = 1.6in]{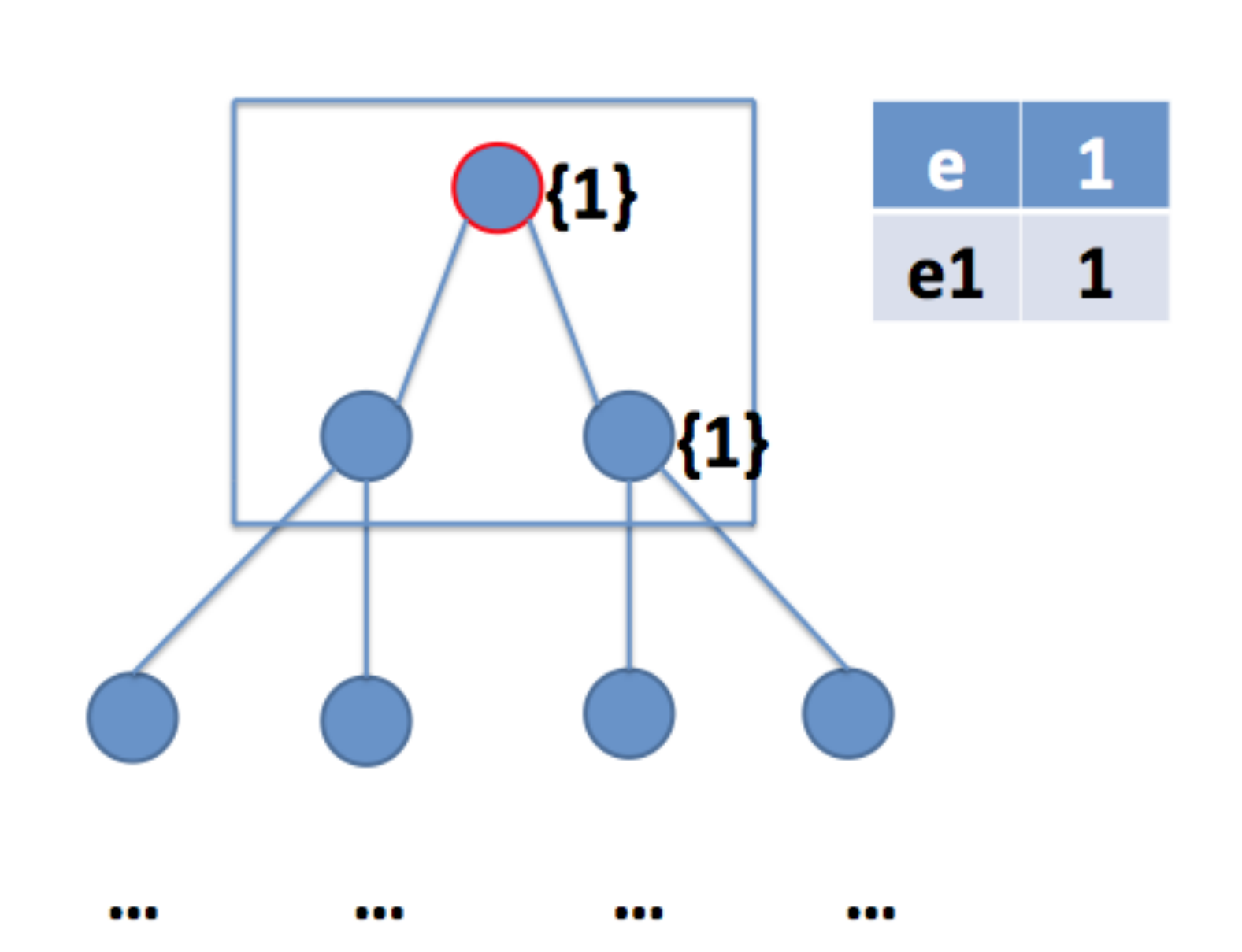}\\
d) \includegraphics[width = 1.6in]{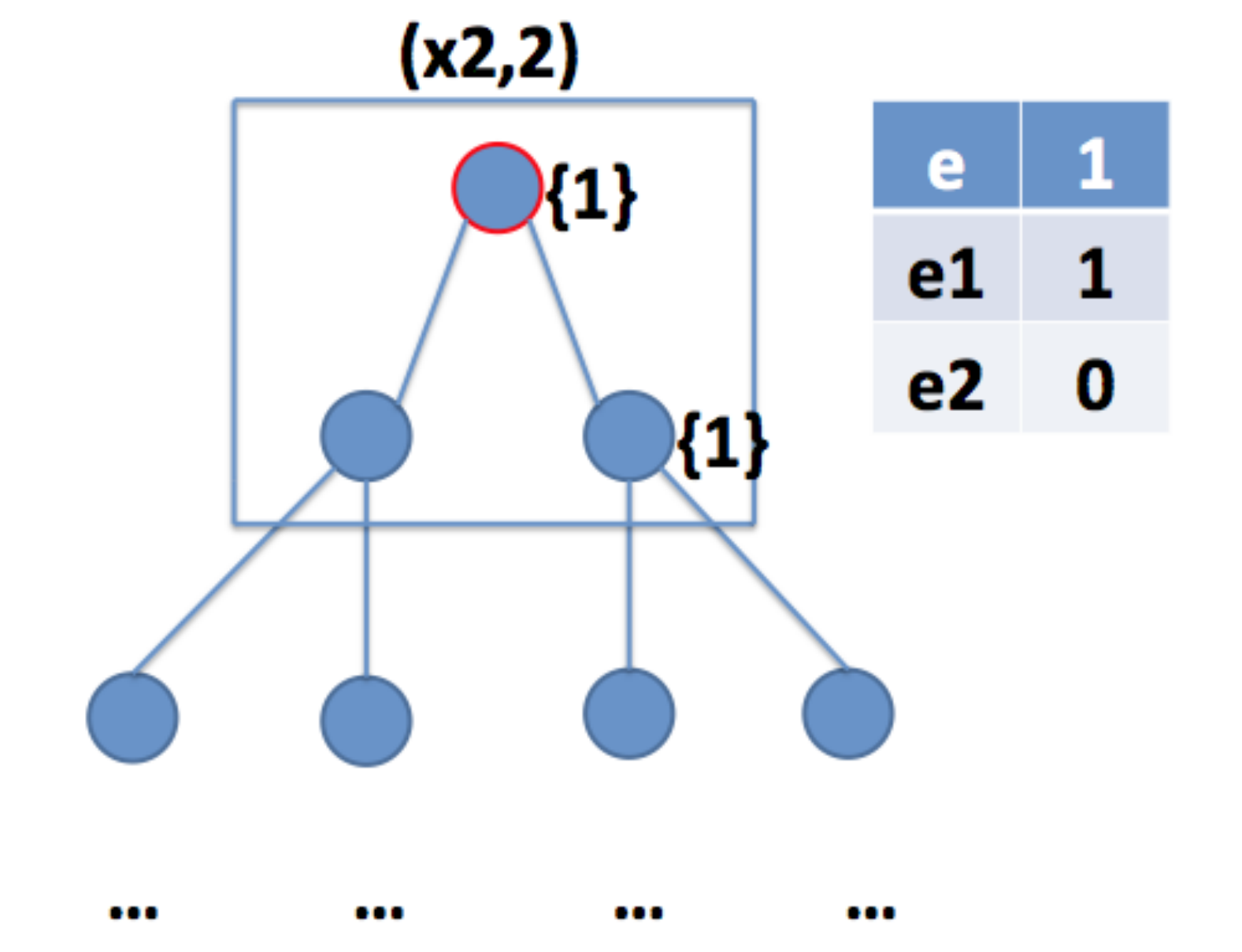}
e) \includegraphics[width = 1.6in]{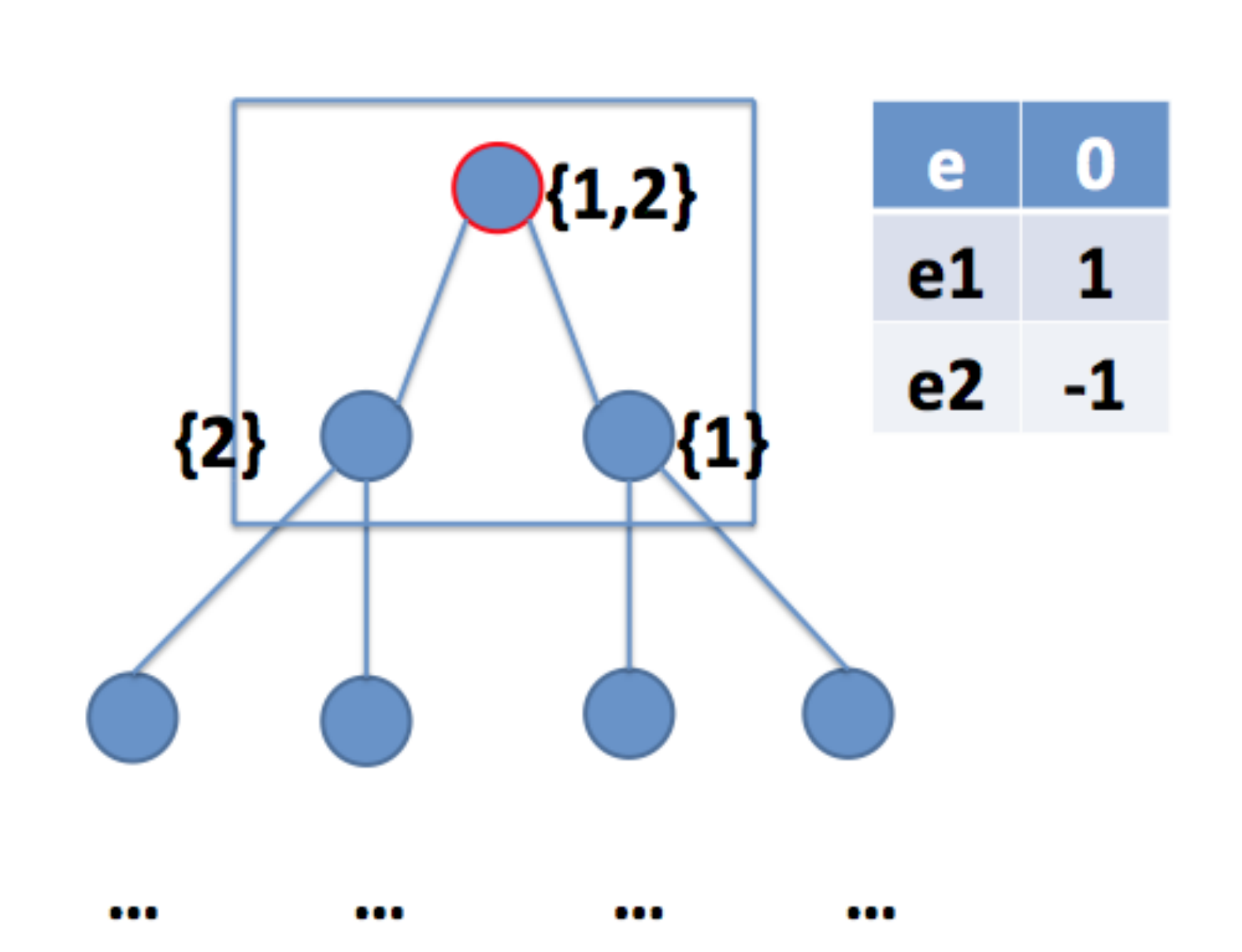}
f) \includegraphics[width = 1.6in]{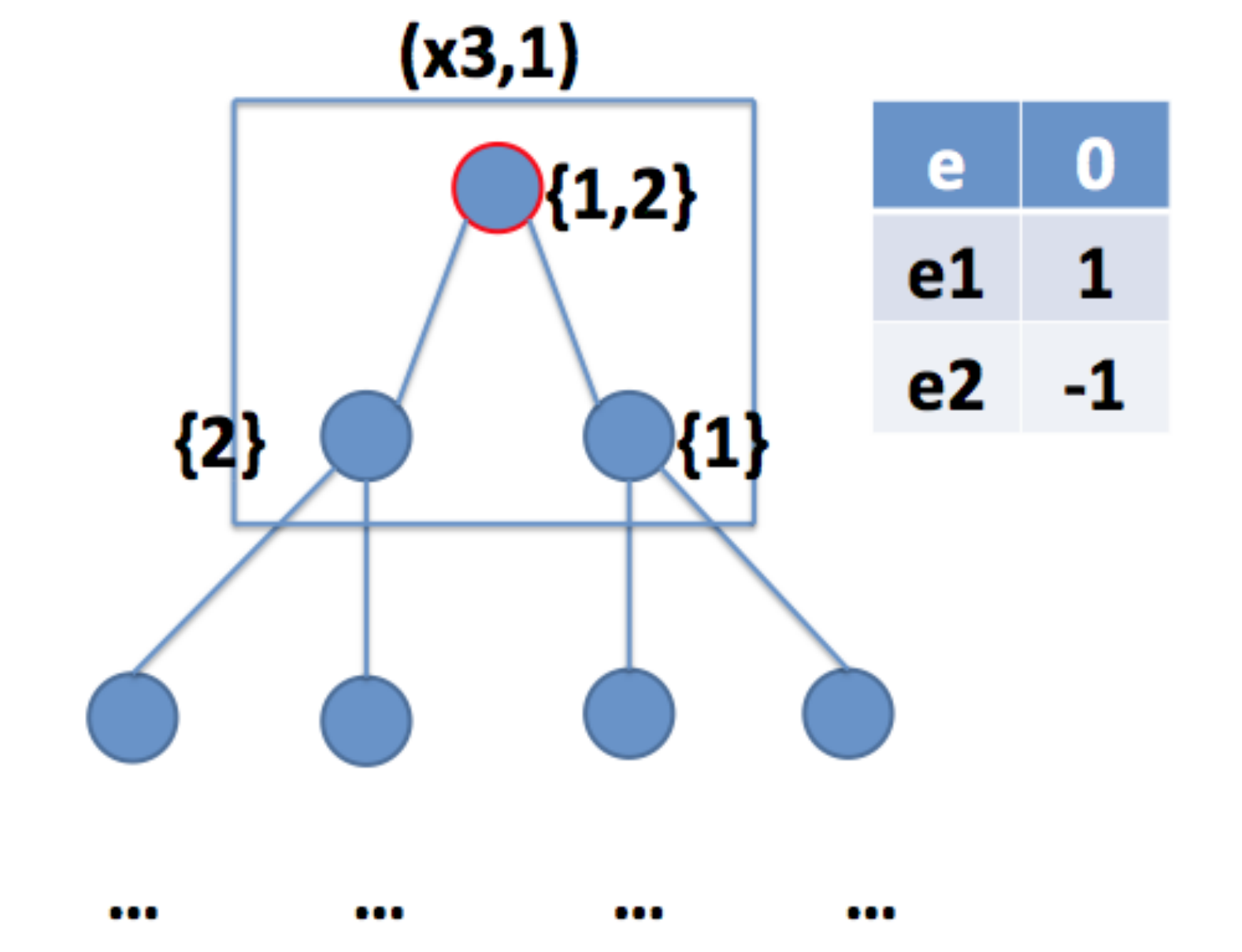}\\
g) \includegraphics[width = 1.6in]{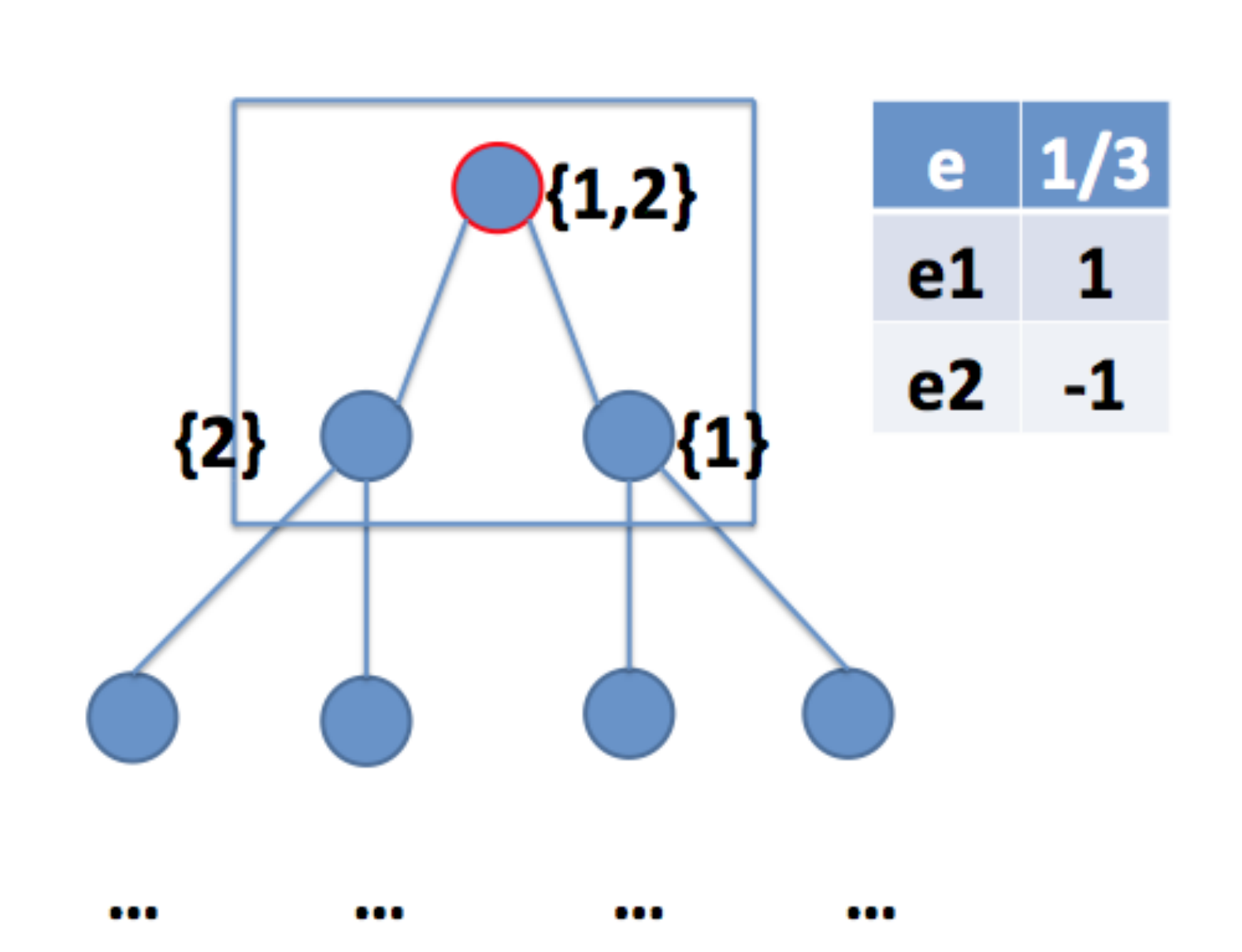}
h) \includegraphics[width = 1.6in]{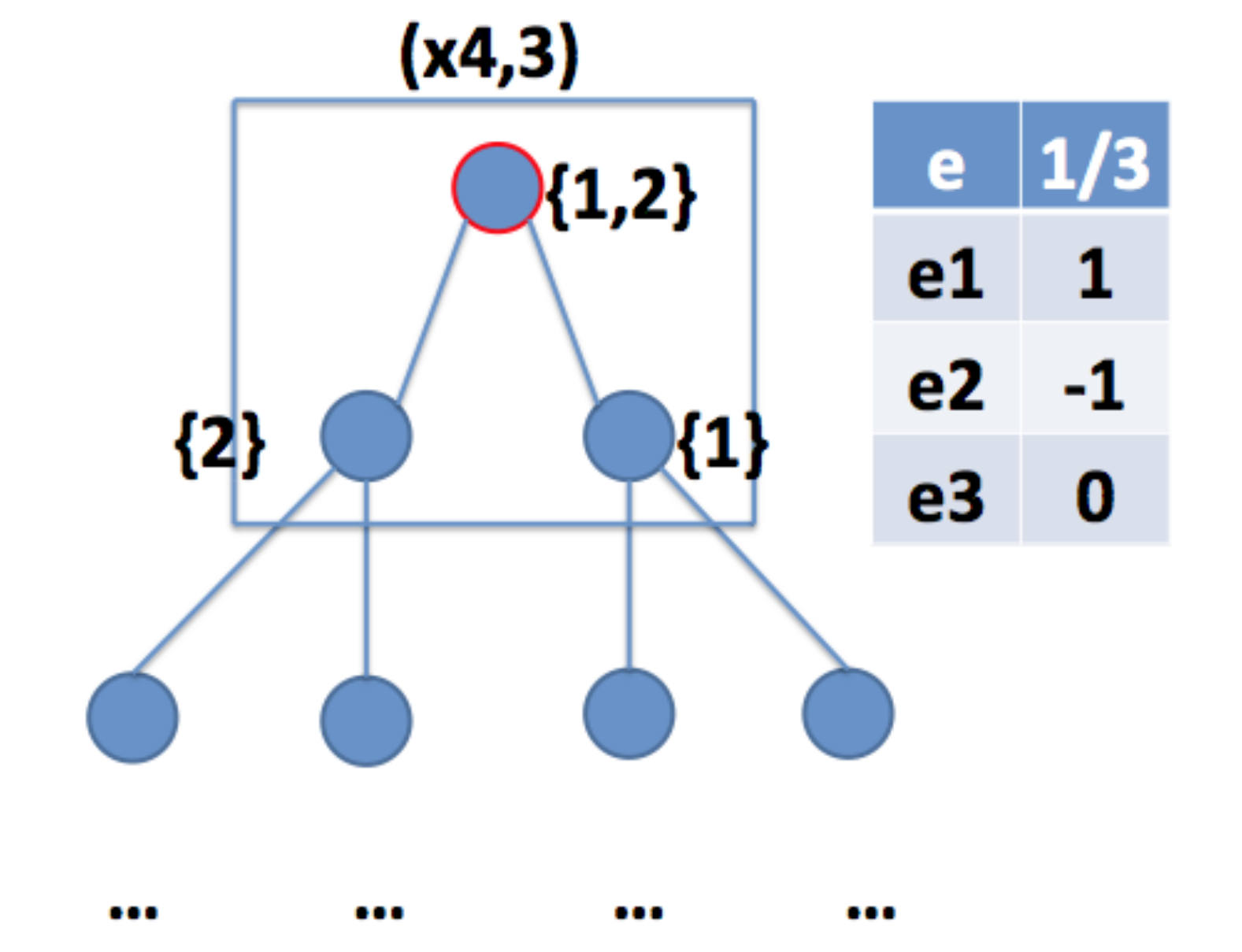}
i) \includegraphics[width = 1.6in]{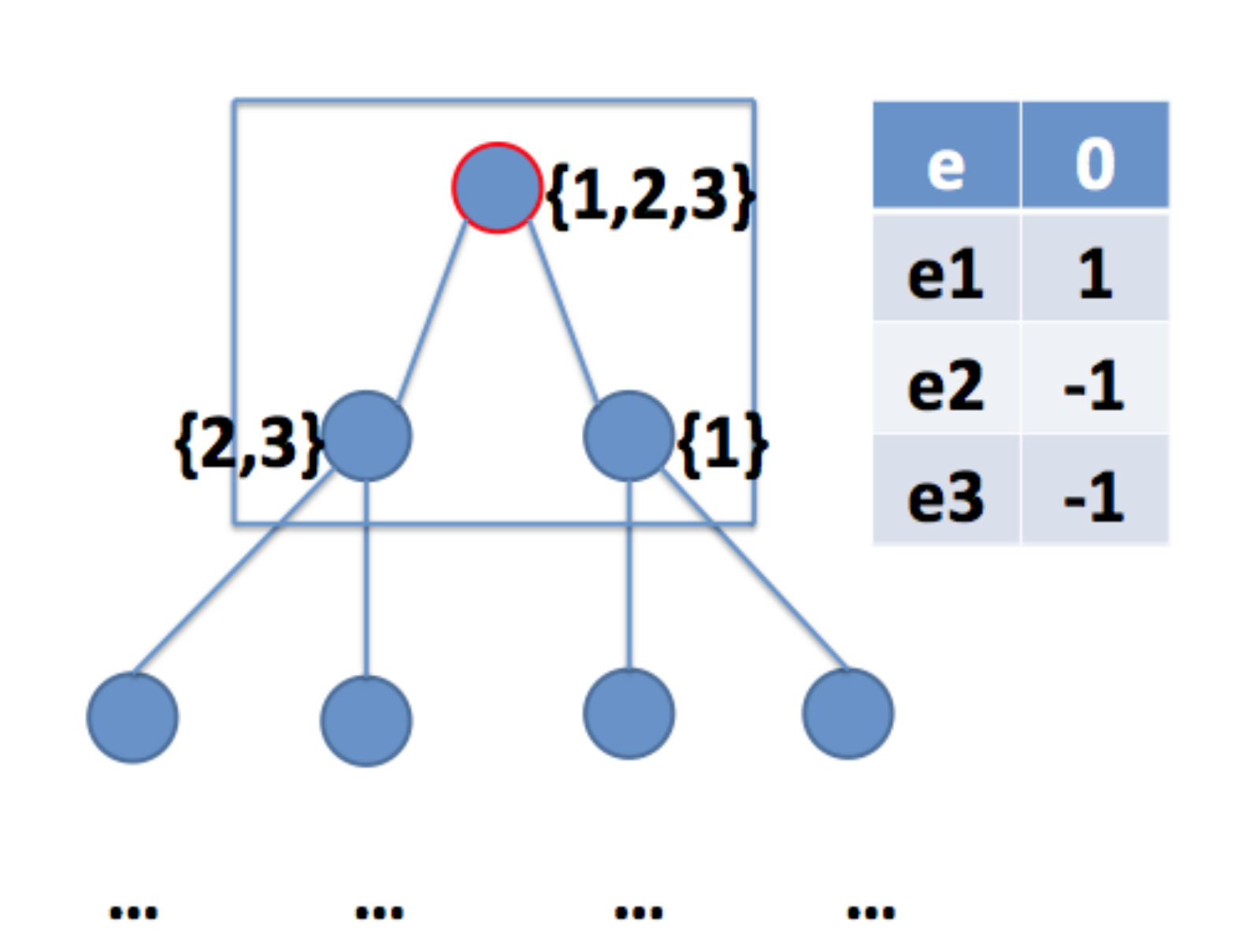}\\
j) \includegraphics[width = 1.6in]{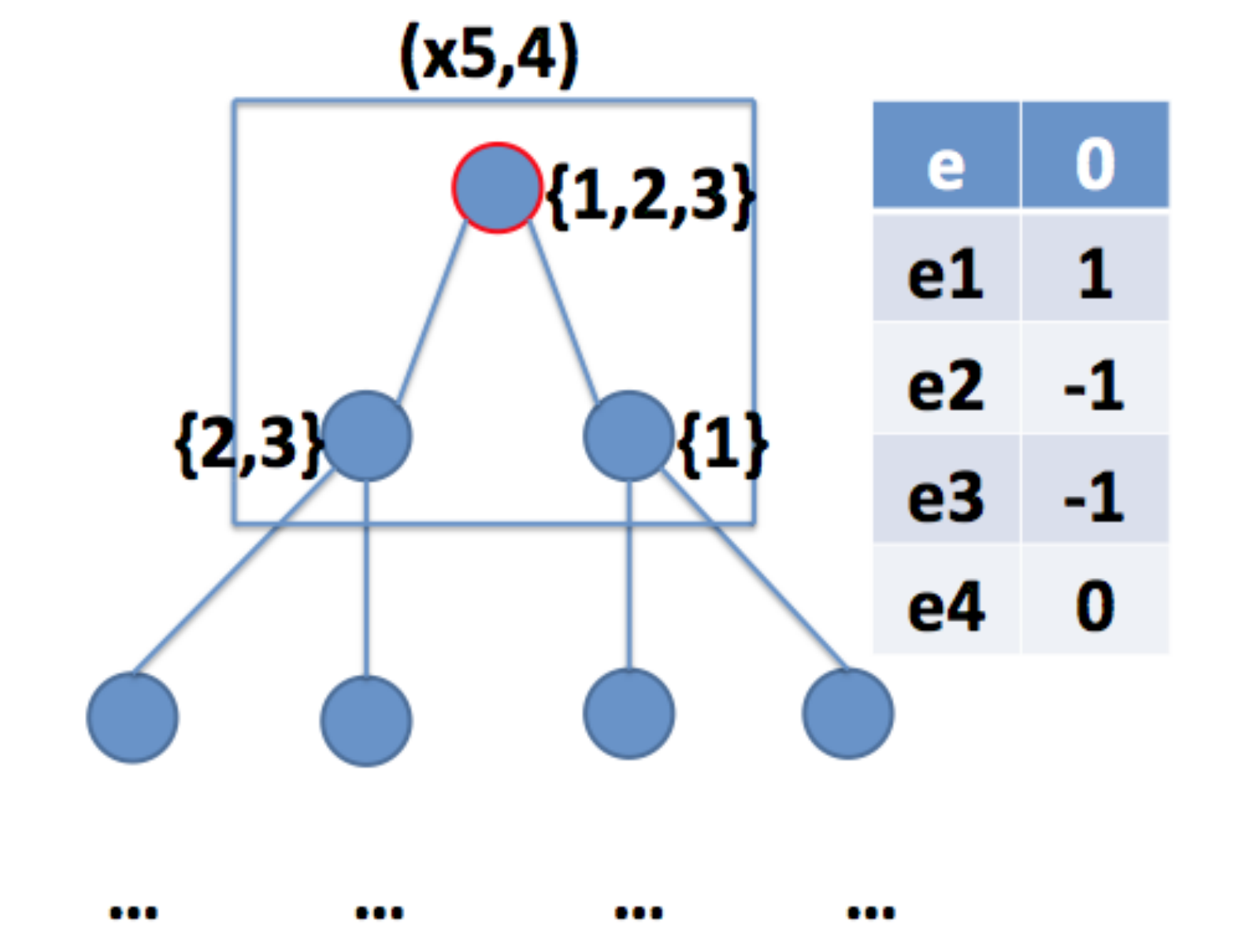}
k) \includegraphics[width = 1.6in]{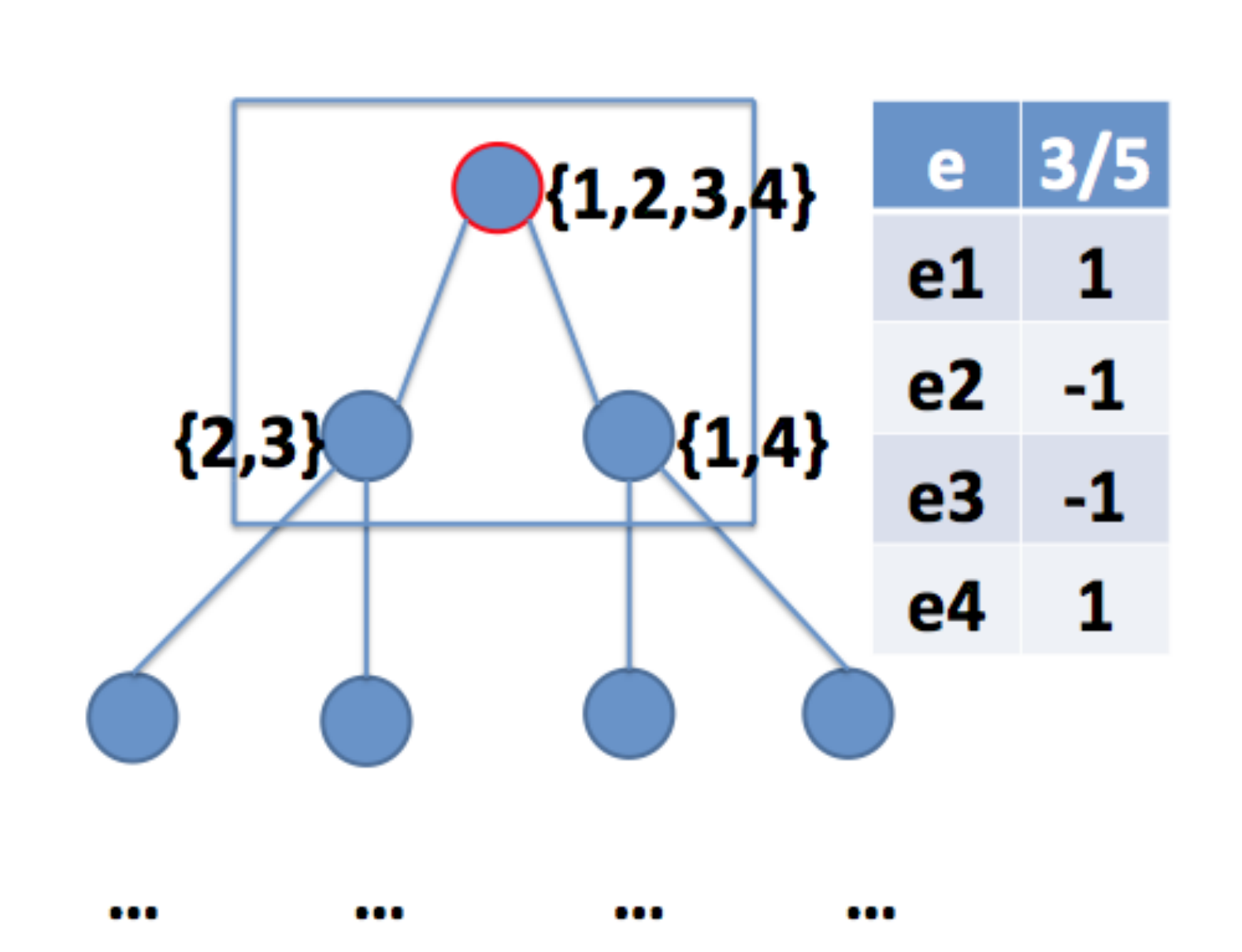}
\vspace{-0.05in}
\caption{Toy example of the behavior of LOMtree algorithm in the tree root.}
\label{fig:root_example}
\end{figure}

Figure~\ref{fig:root_example} shows the toy example of the behavior of LOMtree algorithm for the first few data points. Without loss of generality we consider the root node (exactly the same actions would be performed in any other tree node). Notice that the algorithm achieves simultaneously balanced and pure split of classes reaching the considered node. 

$e$ denotes the expectation $\mathbb{E}_x[h(x)]$, and $e1,e2,e3,e4$ denote the expectations $\mathbb{E}_x[h(x)|i = 1]$, $\mathbb{E}_x[h(x)|i = 2]$, $\mathbb{E}_x[h(x)|i = 3]$, and $\mathbb{E}_x[h(x)|i = 4]$. For simplicity we assume score $h(x)$ can only be either $1$ (if the example is sent to the right) or $-1$ (if the example is sent to the left). The figure should be read as follows (we explain how to read first few illustrations):
\vspace{-0.1in}
\begin{enumerate}[label=\alph*)]
\item Root is initialized. Expectation $e$ is initialized to $0$.
\vspace{-0.05in}
\item The first example $x1$ comes with label $1$ (we denote it as $(x1,1)$). $e1$ is initialized to $0$. The difference between $e$ and $e1$ is computed: $e - e1 = 0$. The difference is non-positive thus the example is sent to the right child of the root, which is now being created (the left child is created along with the right child as we always create both children of any node simultaneously).
\vspace{-0.05in}
\item Expectations $e$ and $e1$ get updated. It is shown that root and its right child saw an example of class $1$.
\vspace{-0.05in}
\item The second example $x2$ comes with label $2$ (we denote it as $(x2,2)$). $e2$ is initialized to $0$. The difference between $e$ and $e2$ is computed: $e - e2 = 1$. The difference is positive thus the example is sent to the left child of the root.
\vspace{-0.05in}
\item Expectations $e$ and $e2$ get updated. It is shown that root saw examples of class $1$ and $2$, whereas its resp. left and right child saw example of class resp. $2$ and $1$.
\vspace{-0.05in}
\item $\dots$
\end{enumerate}
\vspace{-0.2in}

\section{Experiments - dataset details}
Below we provide the details of the datasets that we were using for the experiments in Section~\ref{sec:experiments}:
\begin{itemize}
\item \textit{Isolet}: downloaded from \url{http://www.cs.huji.ac.il/~shais/datasets/ClassificationDatasets.html}
\item \textit{Sector} and \textit{Aloi}: downloaded from \url{http://www.csie.ntu.edu.tw/~cjlin/libsvmtools/datasets/multiclass.html}
\item \textit{ImageNet}~\cite{Deng09imagenet:a}: features extracted according to \url{http://www.di.ens.fr/willow/research/cnn/}, dataset obtained from the authors.
\item \textit{ODP}~\cite{conf/sigir/BennettN09}: obtained from Paul Bennett.  Our version has significantly more classes than reported in the cited paper because we use the entire dataset.
\end{itemize}

\end{document}